\documentclass[letterpaper, 10 pt, conference]{ieeeconf}
\IEEEoverridecommandlockouts
\usepackage{amsmath}
\usepackage{subcaption}
\usepackage{booktabs}
\usepackage{cite}
\usepackage{amsmath,amssymb,amsfonts}
\usepackage{graphicx}
\usepackage{textcomp}
\usepackage[linesnumbered,ruled,vlined]{algorithm2e}
\usepackage{float}
\usepackage{nccmath}
\usepackage[table,xcdraw]{xcolor}
\usepackage{soul}
\usepackage{color, xcolor}

\title{\LARGE \bf Beyond Visibility Limits: A DRL-Based Navigation Strategy for Unexpected Obstacles
}
\author{Mingao Tan$^{1\dagger}$, Shanze Wang$^{1, 2\dagger}$, Biao Huang$^{1,4}$, Zhibo Yang$^{3}$, Rongfei Chen$^{1}$, \\Xiaoyu Shen$^{1}$ and Wei Zhang$^{1}$ 
\thanks{This work has been submitted to the IEEE for possible publication. Copyright may be transferred without notice, after which this version may no longer be accessible.}
\thanks{*This work is supported by 2035 Key Research and Development Program of Ningbo City under Grant No.2024Z127.\textit{(Corresponding author: Wei Zhang.)} }
\thanks{$^{1} $Mingao Tan, Shanze Wang, Biao Huang, Rongfei Chen, Xiaoyu Shen and Wei Zhang are with the Ningbo Key Laboratory of Spatial Intelligence and Digital Derivative, Institute of Digital Twin, Eastern Institute of Technology, Ningbo, China,
        {\tt\small mtan@eitech.edu.cn, szwang@eitech.edu.cn, 210810114@stu.hit.edu.cn, rfchen@eitech.edu.cn, xyshen@eitech.edu.cn, zhw@eitech.edu.cn; }}%
\thanks{$^{2}$Shanze Wang and Hailong Huang are with the Department of Aeronautical and Aviation Engineering, Hong Kong Polytechnic University,{\tt\small shanze.wang@connect.polyu.hk}}%
\thanks{$^{3}$Zhibo Yang is with the Department of Mechanical Engineering, National University of Singapore,{\tt\small zhibo.yang@u.nus.edu}}%
\thanks{$^{4}$Biao Huang is with the School of Science, Harbin Institute of Technology, Shenzhen,{\tt\small 210810114@stu.hit.edu.cn}}%
\thanks{$^{\dagger}$These authors contributed equally to this work as co-first authors.}}

\begin{document}
\maketitle
\thispagestyle{empty}
\pagestyle{empty}

\begin{abstract}

Distance-based reward mechanisms in deep reinforcement learning (DRL) navigation systems suffer from critical safety limitations in dynamic environments, frequently resulting in collisions when visibility is restricted. We propose DRL-NSUO, a novel navigation strategy for unexpected obstacles that leverages the rate of change in LiDAR data as a dynamic environmental perception element. Our approach incorporates a composite reward function with environmental change rate constraints and dynamically adjusted weights through curriculum learning, enabling robots to autonomously balance between path efficiency and safety maximization. We enhance sensitivity to nearby obstacles by implementing short-range feature preprocessing of LiDAR data. Experimental results demonstrate that this method significantly improves both robot and pedestrian safety in complex scenarios compared to traditional DRL-based methods. When evaluated on the BARN navigation dataset, our method achieved superior performance with success rates of 94.0\% at 0.5 m/s and 91.0\% at 1.0 m/s, outperforming conservative obstacle expansion strategies. These results validate DRL-NSUO's enhanced practicality and safety for human-robot collaborative environments, including intelligent logistics applications.


\end{abstract}

\section{Introduction}
Deep Reinforcement Learning (DRL) has emerged as a promising approach for navigation in dynamic environments\cite{Mnih2015HumanlevelCT}. DRL-based methods enable autonomous learning of optimal strategies through environmental interaction, demonstrating robust adaptability. However, existing research primarily emphasizes path efficiency and obstacle avoidance while overlooking a critical factor: the rate of environmental change and its impact on navigation safety\cite{af5ba4575e6c494380b43fc4d7e7b49d}.
Traditional navigation algorithms prioritize path optimality, computing shortest or energy-efficient trajectories. Similarly, DRL-based approaches typically employ distance-based rewards (Euclidean distance to target), where reward magnitude correlates with proximity to the goal\cite{9317723}. While this strategy mitigates sparse reward challenges, it exhibits significant limitations in complex, dynamic environments. In narrow corridors, for instance, robots tend to select paths adjacent to obstacles, creating insufficient observation space for unexpected entities at corners and increasing collision risk.
This limitation stems from neglecting dynamic environmental characteristics, the rate of environmental change, which measures the speed and magnitude of variations near the robot during movement. In closed-corner turns, minimizing environmental changes provides three critical benefits: expanded observation space, optimized turning speeds, and improved obstacle avoidance capability. Consequently, ignoring environmental change rates leads to delayed or suboptimal decisions that compromise navigation safety\cite{cimurs2021goal}.

To address these challenges, we propose a novel DRL-based Navigation Strategy for Unexpected Obstacles (DRL-NSUO) that ensures safer and more reliable autonomous navigation. Our method incorporates environmental change rates into the reinforcement learning reward mechanism through real-time LiDAR data monitoring and dynamic strategy adjustment. The reward function integrates path efficiency objectives with environmental change rate constraints, while curriculum learning progressively enhances change rate penalties, strengthening the robot's decision-making in dynamic environments. Additionally, we improve obstacle perception through LiDAR data preprocessing, emphasizing short-range scan data for precise environmental dynamic detection.
The key contributions of this study include:
\begin{itemize}
\item Development of DRL-NSUO, a method that significantly enhances navigation safety compared to traditional DRL-based approaches while maintaining superior obstacle avoidance capabilities in crowded environments.
\item Comprehensive validation in the BARN dataset\cite{9292572} demonstrating substantial performance improvements over traditional and advanced learning-based approaches, with success rates of 94.0\% at 0.5 m/s and 91.0\% at 1.0 m/s.
\item Extensive real-world experiments validating effective navigation performance and exceptional safety capabilities in complex static and dynamic environments, ensuring the safety of both robots and pedestrians.
\end{itemize}

\section{Related Work}




\subsection{Traditional Local Planners}
Traditional navigation methods prioritize operational reliability in structured environments. The Dynamic Window Approach (DWA)\cite{Fox1997TheDW} implements dynamic velocity adjustment using real-time obstacle data; however, its limited planning horizon renders it ineffective in densely crowded environments. Artificial Potential Field (APF)\cite{Khatib1985RealTimeOA} methods utilize repulsive force representations for obstacles; nevertheless, these methods consistently encounter local minima problems. The Vector Field Histogram (VFH)\cite{Borenstein1991TheVF} method addresses these limitations through path selection based on minimal resistance in grid-based obstacle maps; however, its dependence on fixed grid resolutions compromises performance in scenarios with fast-moving obstacles. The Timed Elastic Band (TEB)\cite{Rosmann2015TEB} approach incorporates kinodynamic constraints and temporal aspects for trajectory optimization; While these methods are computationally efficient and effective in structured environments, they often struggle in dynamic, cluttered and crowded settings where global context or anticipatory behavior is required.

\subsection{DRL-Based Navigation Methods}

Deep Reinforcement Learning (DRL) has transformed robotic navigation by facilitating the acquisition of complex policies in high-dimensional state spaces. Contemporary research focuses on improving navigation efficiency through reward engineering. Studies by \cite{Choi2020Fast,Miranda2022Generalization} employ distance-based rewards to minimize path length; however, their inflexible reward structures do not account for dynamic obstacles interactions and environmental volatility. Hierarchical reinforcement learning approaches \cite{Lee2023Adaptive} implement layered decision-making through meta-policy training for low-level skill switching, achieving enhanced adaptability at the expense of real-time obstacle responsiveness. Adversarial RL frameworks \cite{Wisniewski2024Benchmarking} strengthen robustness against sensor noise through domain randomization, while model-based RL \cite{Zhu2023A} optimizes sample efficiency by integrating environment dynamics; nevertheless, both approaches lack explicit mechanisms to manage occlusions and sudden environmental changes. Research by \cite{Zhuang2024Progress} advances sensor-denied navigation through adversarial training; however, the persistence of static goal-centric rewards constrains adaptability in cluttered environments.

\begin{figure}[t]
	\centering
	\includegraphics[width=0.45\linewidth]{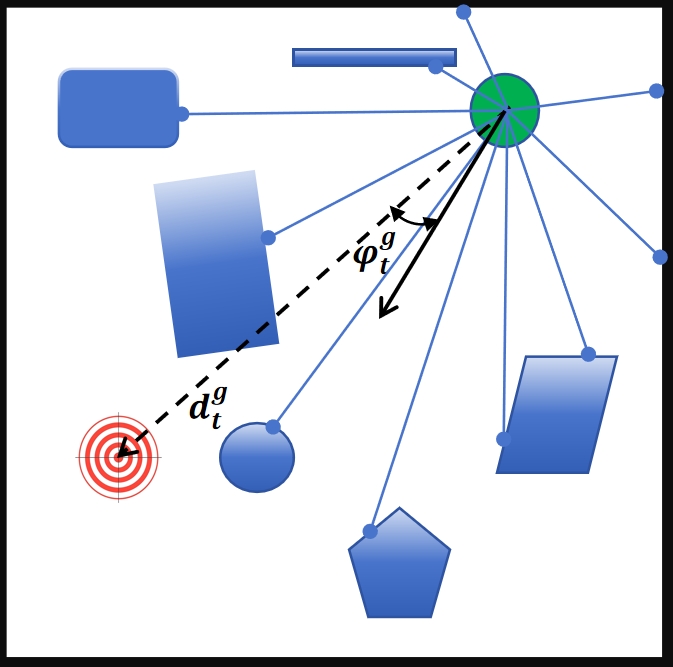}
	\caption{Illustration of the goal-driven robot navigation problem.}
	\label{fig1}
\end{figure}

\section{Preliminaries}

\subsection{Problem Formulation}

\begin{figure}[t]
    \centering
    \includegraphics[width=0.5\linewidth]{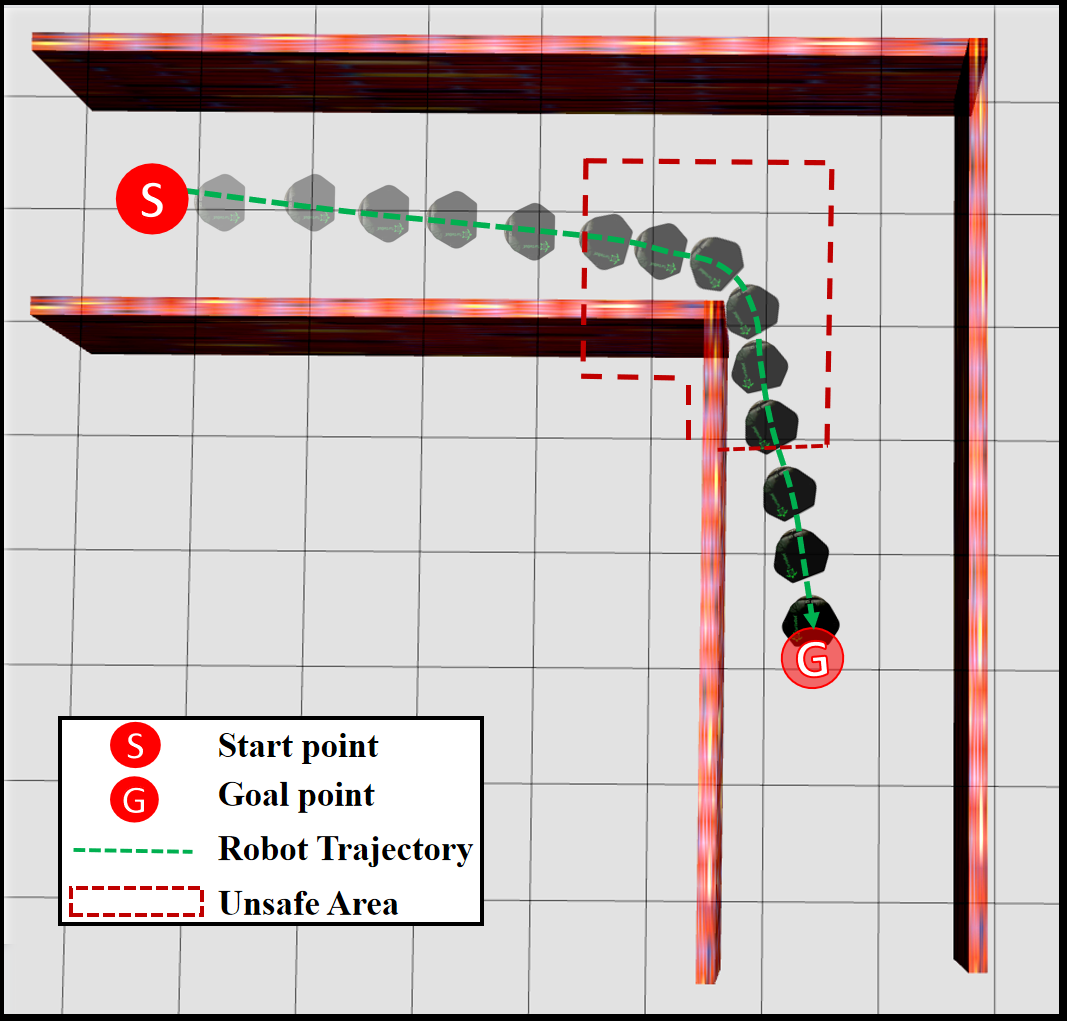}
    \caption{The diagram demonstrates the distance-based robot's behavior of maintaining proximity to the wall during its traversal. In real-world test, it may result in collisions or pedestrian injuries, as shown in Fig. 10b and Fig. 10d.}
    \label{figure1}
\end{figure}

The problem is formulated as a sequential decision-making process. As shown in Fig. \ref{fig1} illustrates a LiDAR-equipped robot navigating toward its goal in a complex, crowded environment while maintaining collision-free operation. The goal's relative position, denoted as $s_t^g =\{d_t^g, \varphi_t^g\}$, is acquired through localization sensors. The control policy $\pi_\theta$ is implemented via a deep neural network (DNN) controller with parameters $\theta$. The controller processes the input state $s_t =\{s_t^g, s_t^o\}$, where $s_t^o$ contains local environmental data acquired from 2D LiDAR scans. The LiDAR sensor's coordinate frame is calibrated to the robot's local reference frame, with its origin positioned between the drive wheels. At each discrete time step $t$, the policy $\pi_\theta$ generates velocity commands $a_t =\{v_t, \omega_t\}$ based on state $s_t$, yielding a reward $r_t^{\text{nav}}$.The reward function of the traditional DRL-based navigation method is:
\begin{equation}
r_t^{\text{nav}}=\left\{\begin{matrix}r_{\text{reach}},&\text{if reaches the goal,}\\r_{\text{crash}},&\text{if collides,}\\c_1\left(d_t^g-d_{t+1}^g\right),&\text{otherwise,}\\\end{matrix}\right.
\end{equation}
where $c_1$ represents a scaling coefficient. In deep reinforcement learning (DRL) for robot navigation, traditional reward functions are typically based on distance, such as $c_1\left(d_t^g-d_{t+1}^g\right)$, where rewards increase as the robot moves closer to the goal. However, these Euclidean distance-based rewards have significant limitations in complex environments, especially those with corners or narrow passages. The reward gradient continuously directs the robot toward the target, which can trap the robot into choosing the shortest path, neglecting obstacles. As shown in Fig. \ref{figure1}, such behavior poses serious risks, as it can lead robots to fail to respond effectively to unknown obstacles, potentially causing harm to pedestrians or damaging the environment.

The objective is to design an additional representation reward $r_t^{\text{env}}$ that can improve the safety performance of robots. By combining $r_t^{\text{env}}$ with $r_t^{\text{nav}}$, the drawbacks of the distance-based reward design can be mitigated. This approach maximizes the robot's ability to navigate in crowded environments while minimizing the rate of change in the surrounding environment during the robot's movement, thus forming a safe observation space that can effectively avoid obstacles.
The goal of the training process is to learn the optimal policy $\pi_\theta^\ast$ that maximizes the expected cumulative reward, $G_t = \sum_{\tau=t}^T \gamma^{\tau-t} r_\tau,$where $\gamma$ is the discount factor. The optimal policy $\pi_\theta^\ast$ must demonstrate effective generalization capabilities, ensuring robust performance across diverse scenarios while maintaining reliability in previously unexpected environments.

\section{Approach}
To address this issue, we propose a mechanism for sensing the rate of environmental variation derived from LiDAR data. We then introduce the neural network architecture and preprocessing methods for LiDAR data. Finally, we integrate the environmental change rate into the reinforcement learning reward mechanism by monitoring LiDAR data variation in real-time, while using curriculum learning to gradually enhance the robot's safety performance.
\subsection{Mechanism for Sensing the Rate of the Environment Variation}
The robot moves at different speeds in different environments, making it challenging to precisely describe the changes in its surroundings. 
To better capture the dynamics of the robot's surrounding environment, we propose a metric based on the real-time environmental change rate derived from LiDAR's real-time measurement data.The adjusted rate of environmental change $v$ at time step $t$ is defined as:
\begin{equation}
v_{c}^{(t)} = \left( \frac{\sum_{i=n_1}^{n_2} \left| s_{\text{scan},i}^{(t+1)} \right|}{\sum_{i=n_1}^{n_2} \left| s_{\text{scan},i}^{(t)} \right|} - c_1 \right) \cdot c_2 + c_1
\label{eq:envrate}
\end{equation}
Here, $s_{\text{scan},i}^{(t+1)}$ represents the absolute value of the $i$-th data point in the current laser scan at time $t+1$, and $s_{\text{scan},i}^{(t)}$ corresponds to the same data point from the previous time step $t$. The range $[n_1, n_2]$ defines the selected indices for the LiDAR’ data. The term $c_1$ is the baseline offset ($c_1 = 1.0$), and $c_2$ is the scaling factor ($c_2 = 10.0$).


\begin{figure}[t]
    \centering
    \includegraphics[width=0.9\linewidth]{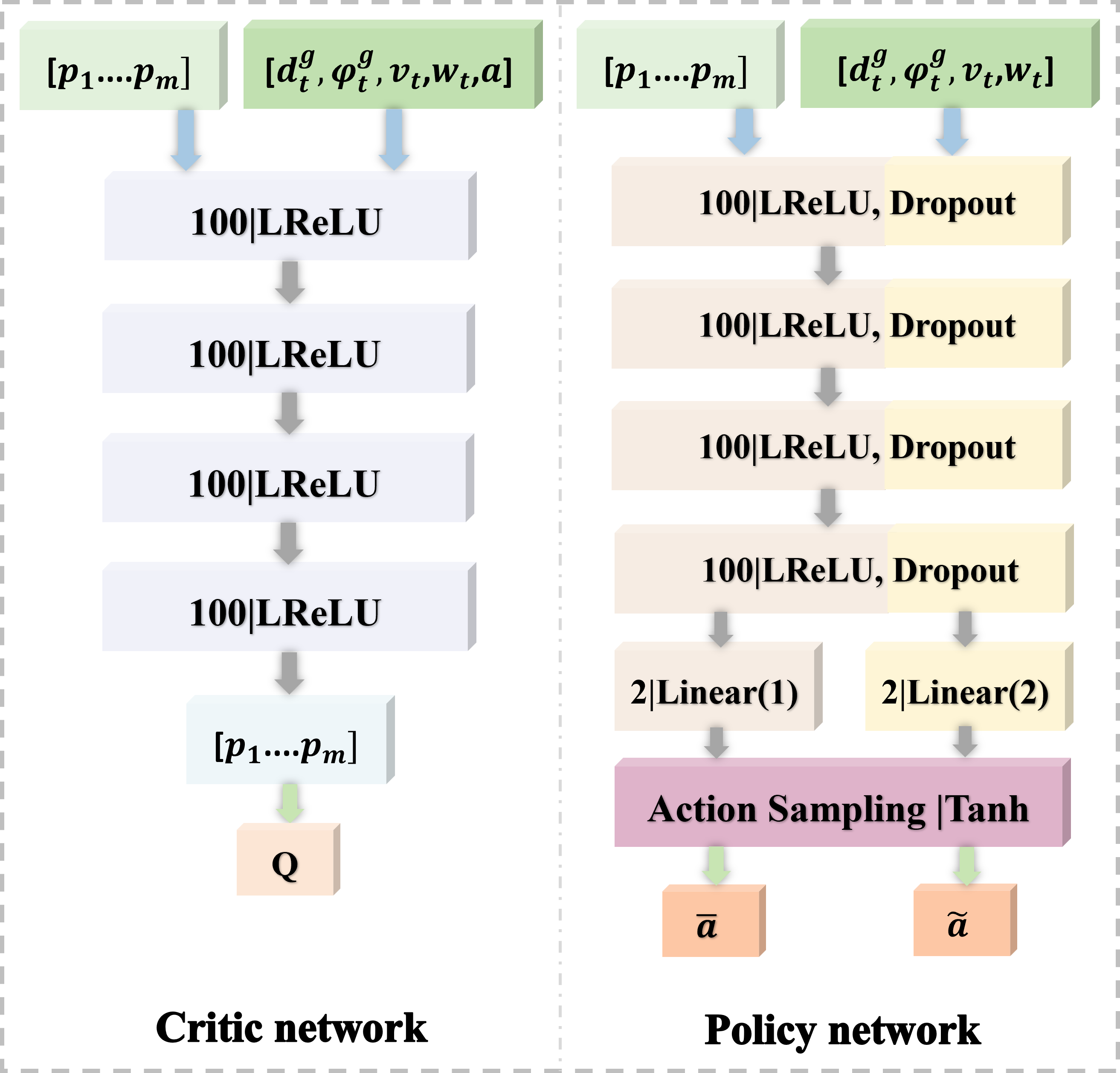}
    \caption{Neural Network architectures used for training.}
    \label{figure3}
\end{figure}

\subsection{Neural network architecture}
This paper uses the SAC algorithm \cite{haarnoja2018soft} for deep reinforcement learning-based navigation. To improve generalization, we employ a four-layer fully-connected (FC) model with dropout-based policy regularization. For computational efficiency, the original laser scan data (1080 beams) is down-sampled to $N=30$ values. The computation for each down-sampled value $o_{\min}^i$ is defined as:

\begin{equation}
\begin{aligned}
d^o_{\mathrm{min},i} = \min(d^o_{i\cdot k+1}, d^o_{i\cdot k+2}, \cdots, d^o_{i\cdot k+k-1})
\end{aligned}
\end{equation}where $i$ denotes the index of the down-sampled laser scans, and $k$ represents the down-sampling window length ($k=36$). The state vector $s$, comprising down-sampled laser scans, relative goal position, and robot velocities, is processed by the policy network and the value network to compute action $a=\pi_\theta(s)$ and value function $V_\psi(s)$, respectively. This implementation employs a squashed Gaussian SAC policy, restricting the actions sampled to the interval $[-1,1]$. During the training process, the sampled action $a_\mathrm{sam}(s|\theta)$ from the squashed Gaussian policy is computed as: $a_\mathrm{sam}(s|\theta)=\tanh{\left(\mu_\theta(s)+\sigma_\theta(s)\odot\zeta\right)\ }$,where $\mu_\theta\left(s\right)$ and $\sigma_\theta(s)$ denote the mean and standard deviation of the Gaussian policy, and $\zeta \sim \mathcal{N}(0,1)$. The specific network structure of this method is shown in Fig. \ref{figure3}, where $[p_1...p_m]$ is preprocessed LiDAR min-pool data $[d^o_{\mathrm{min},1} ... d^o_{\mathrm{min},m}]$. To enhance the agent's sensitivity to nearby obstacles, the LiDAR data should magnify shorter distances for easier policy learning.
Following the approach in \cite{9804689}, we implement a reciprocal function $p_i^R(\cdot)$ as the inverse perception function:
\begin{equation}
p_i^R = \frac{1}{d^o_{\mathrm{min},i}  - \beta_i}
\end{equation}
where $p_i^R$ represents the preprocessed value of $y_i$, and $\beta_i$ denotes the trainable parameter.

\subsection{A DRL-Based Navigation Strategy for Unexpected Obstacles}

\begin{algorithm}[t]
\caption{Training of DRL-NSUO}
\SetAlgoLined
Initial policy network parameters (with trainable IP parameters) $\boldsymbol{\theta}$, critic network parameters (with trainable IP parameters) $\boldsymbol{\phi}_1$, $\boldsymbol{\phi}_2$, empty replay buffer $\mathcal{B}$, initial learning factor $c = 1.5$\;
\For{\upshape{episode}$=1$, 2, \ldots}{
    Reset the training environment and initialize $t = 0$\;
    Obtain initial observation $x_0$\;
    \While{$t < T_{\mathrm{max}}$ \textbf{and} not terminate}{
        \eIf{training}{
            Sample action $a_t \sim \pi_{\boldsymbol{\theta}}(x_t)$\;
        }
        {
            Sample a random action $a_t$\;
        }
        Execute $a_t$ in simulation\;
        Obtain next observation $x_{t+1}$, navigation reward $r_t^{\text{nav}}$, environment reward $r_t^{\text{env}}$ and termination signal $d_t$\;
        
        \eIf{$c = 1.5$}{
            Obtain speed reward $r_t^{\text{speed}}$\;
        }
        {
            $r_t^{\text{speed}} = 0$\;
        }
        Combine all rewards: $r_t^{\text{all}} \gets r_t^{\text{nav}} + r_t^{\text{env}} + r_t^{\text{speed}}$\;
        Update success rate $\rho$\;
        \If{$\rho_{T} \leq \rho$}{
            $\rho \gets 0$\;
            $c \gets c + 0.5$\;
        }
        
        Store $\{x_t, a_t, r_t^{\text{all}}, x_{t+1}, d_t\}$ in $\mathcal{B}$\;
        $t \gets t + 1$, $x_t \gets x_{t+1}$\;
        \If{training}{
            Sample a mini-batch $\mathcal{M}$ from $\mathcal{B}$\;
            Update $\boldsymbol{\phi}_1$, $\boldsymbol{\phi}_2$\;
            Update $\boldsymbol{\theta}$\;
        }
    }
}
\label{algorithm}
\end{algorithm}

Based on the definition of environmental change rate $v_{c}^{(t)}$ in (\ref{eq:envrate}), we incorporate the environmental change rate into the reinforcement learning reward mechanism by monitoring the variation rate of LiDAR data in real-time. In the reward design, we adopt a curriculum learning approach to gradually increase the reward weight associated with the environmental change rate. This allows the robot to balance rapid obstacle avoidance, extensive exploration capabilities, and minimal environmental changes, ensuring an optimal safety performance navigation algorithm. Here, $r_t^{\text{env}}$ is designed to satisfy the following conditions: When the environmental change rate $v_{c}^{(t)}$ is equal to or near 1, a positive reward is given, indicating that the environmental change rate around the robot during motion is relatively low. Conversely, as $v_{c}^{(t)}$ deviates further from 1, a higher penalty is imposed. Based on this principle, in this experiment, we use a symmetric first-order function to represent the reward function of the rate of change, $r_t^{\text{env}}$ can be expressed as:

\begin{equation}
r_t^{\text{env}} =
\begin{cases} 
    c(\frac{k_1}{v_{c}^{(t)}} - k_2), & \text{if } v_{c}^{(t)} > 1, \\
c(\frac{k_1}{k_1 - v_{c}^{(t)}} - k_2), & \text{if } v_{c}^{(t)} \leq 1.
\end{cases}
\end{equation}
where $c$ is the learning factors for curriculum learning and and $k_1 = 2.0$, $k_2 = 1.9$ are the constant parameters for calculation.The initial value of $c$ is set to 1.5. When the robot's navigation success rate in the current learning factor exceeds the pre-defined threshold $\rho_{T}$ ($\rho_{T}$ = 0.9 in this paper), the value of $c$ is increased by 0.5, and the robot enters the next curriculum. This means that the penalty for drastic environmental changes is increased. By continuously increasing the value of $c$, the robot gradually becomes more sensitive to the rate of environmental changes, thereby enhancing its safety performance in complex environments. Additionally, we select the network that was the last to reach the switching point of the $c$ value. In the specific experiment, we attempted to introduce an additional speed-based reward mechanism during the initial curriculum learning stage (with parameter \(c=1.5\)): $r_t^{\text{all}} = r_t^{\text{nav}} + r_t^{\text{env}} + r_t^{\text{speed}} \cdot \mathbb{I}(c=1.5)$.Once the training success rate first reached the target success rate, we immediately removed this reward component. This design only aims to accelerate the robot's learning speed during the early stages of training. The faster speed behavior exhibited by the robot due to the initial reward mechanism $r_t^{\text{speed}} \cdot \mathbb{I}(c=1.5)$ will be suppressed later by the reward mechanism \( r_t^{\text{env}} \), and the algorithm for training DRL-NSUO is depicted in Algorithm 1, $r_t^{\text{speed}}$ is defined as ($ \beta=0.5 $) :

\begin{equation}
r_{t}^{\text{speed}} = 
\begin{cases} 
\beta \cdot v_t, & \text{if } c = 1.5 \\
0, & \text{otherwise}
\end{cases}
\end{equation}

\begin{figure}[t]
    \centering
    \includegraphics[width=0.4\linewidth]{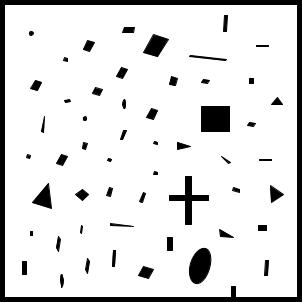}
    \caption{Simulated training environment Env1.}
    \label{fig:4}
\end{figure}

\section{Robot training and test in Simulation}
\subsection{Training and Test in Ros Stage}


The DRL training platform used in this experiment is built on Ros Stage \cite{StageRos}, a widely used lightweight simulator for mobile robots. Each training and testing environment consists of a 2-D map, with the robot controller operating at a frequency of 5 Hz. The maximum time step $T_{\mathrm{max}}$ for each episode is set to 200. As shown in Fig. \ref{fig:4}, the training environment Env1 is a $10 \times 10$  $m^2$ map. The total number of steps trained is 200,000. The experimental platform utilizes a simulated black circular robot with differential drive capability, which is deployed on this map for training to quickly and safely reach the target location. The robot's maximum linear velocity and angular velocity are 0.5 m/s and $\pi/2$ rad/s, respectively.A 2D LiDAR is mounted on the robot for obstacle detection. The LiDAR's field of view (FOV), angular resolution, and maximum sensing range are 270 °, 0.25 ° (1800 laser beams) and 30 meters, respectively. The proposed algorithm implements min-pooling to reduce the laser scan dimensionality to 36. At the beginning of each episode, the agent starts from a random position on the map, aiming to reach a randomly appearing target point. During this process, the agent must avoid both time limit violations and obstacle collisions. To enhance obstacle avoidance capabilities in crowded environments, the challenging environment Env1 is specifically designed to facilitate the development of robust safety strategies and efficient obstacle avoidance behaviors.


\begin{figure}
    \centering
    \includegraphics[width=0.95\linewidth]{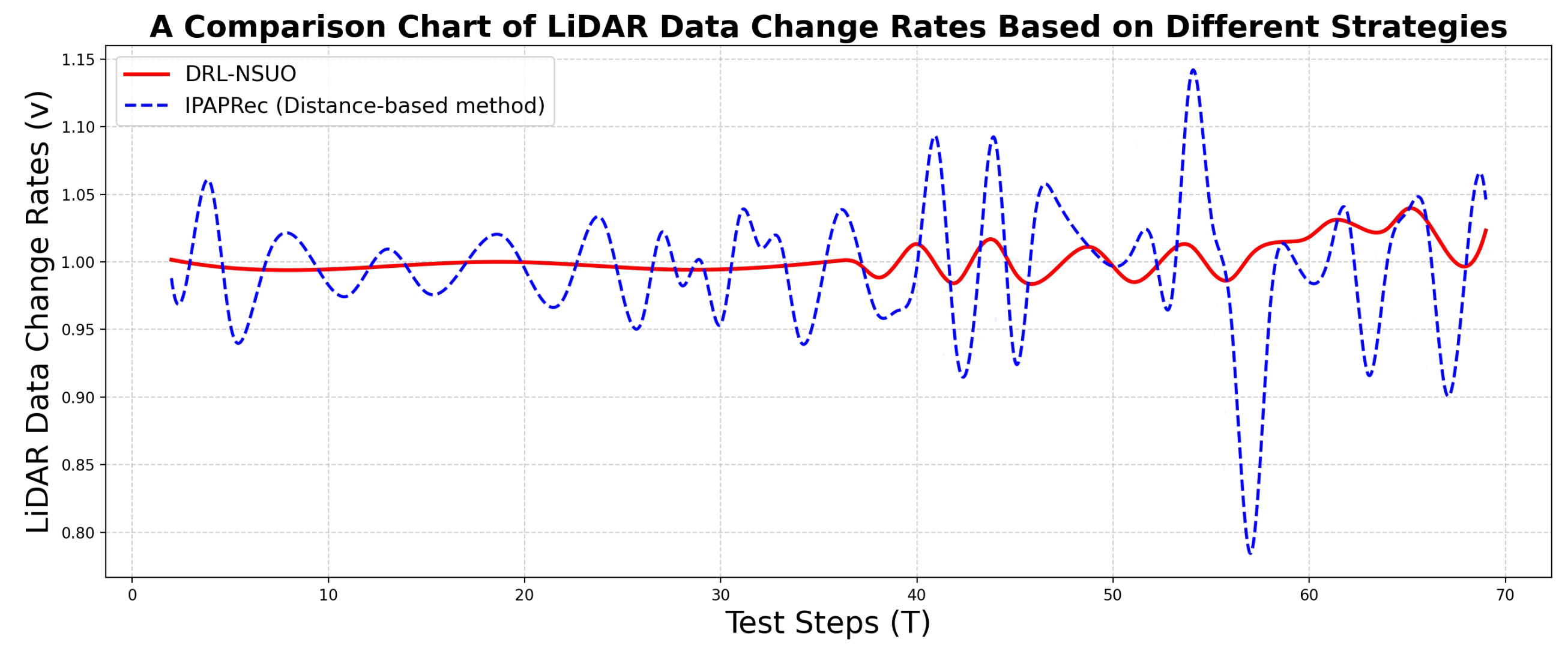}
    \caption{
A comparison chart of the LiDAR data change rate in the environment for DRL-NSUO and IPAPRec (Distance-based DRL algorithm) when passing through the same corner map Env2 (with the same starting point, endpoint, speed, and acceleration).}
    \label{comparison chart}
\end{figure}
\begin{figure}[t]
	\centering
	\subfloat[DRL-NSUO]{%
		\includegraphics[width=0.49\linewidth]{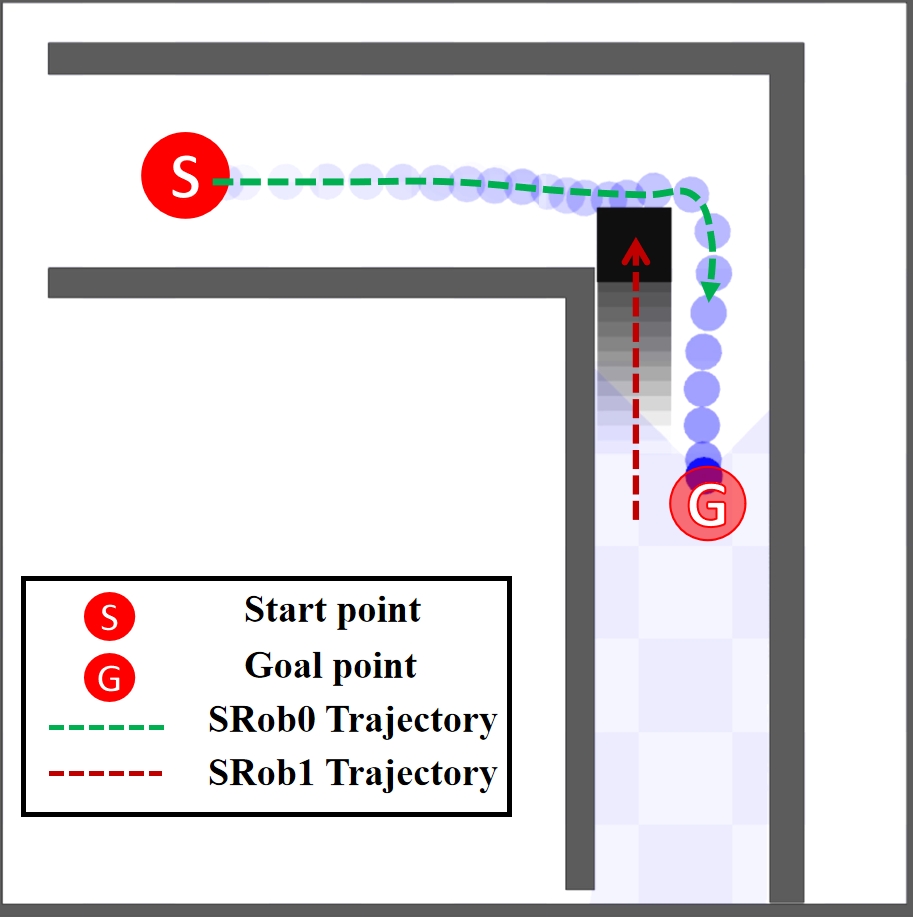}
		\label{chutian1}
	} 
	\subfloat[IPAPRec(distance-based)]{%
		\includegraphics[width=0.49\linewidth]{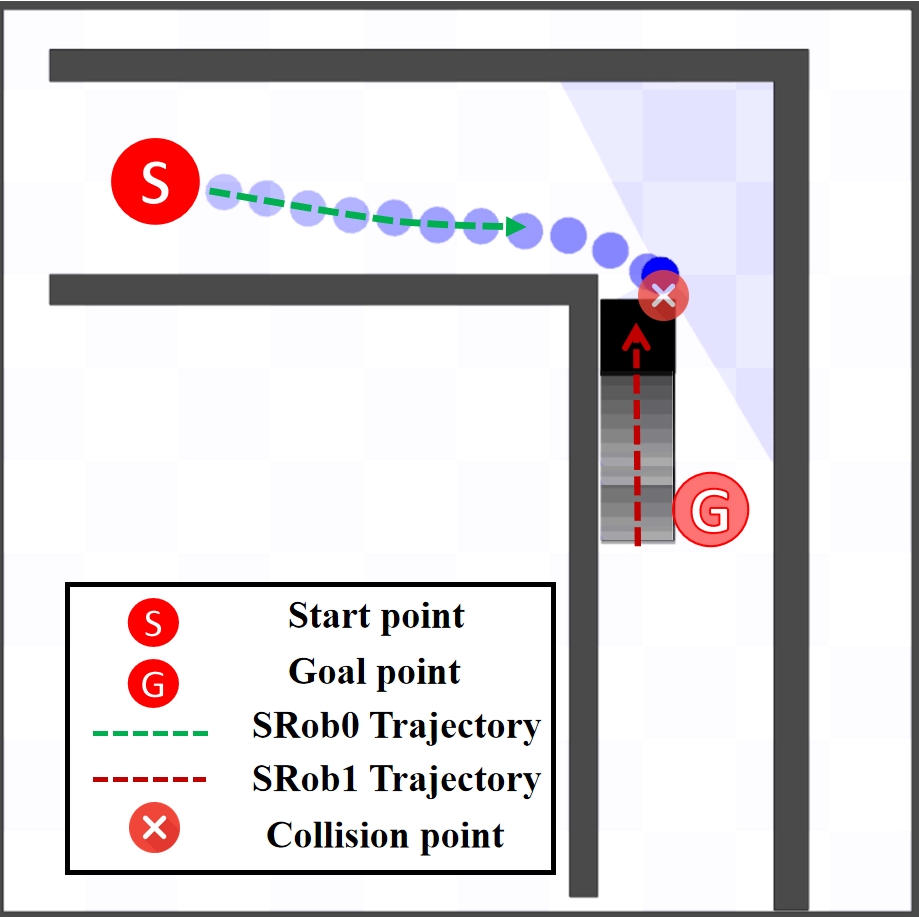}
		\label{chutian2}
	}
	\caption{(a):SRob0 autonomously decelerated at the corner and maintained a safe LiDAR observation space, resulting in a safe distance from SRob1 and successfully avoiding a collision.(b):SRob0 adopts a locally optimal strategy during the turning process and maintains its current speed while turning, which ultimately leads to a collision with SRob1.}
	\label{fig:combined}
\end{figure}

Next, we focus on testing its safety performance(characterized by minimizing the environmental variation rate and maintaining a safe perspective for rapid obstacle avoidance) in both the Stage simulation environment and real-world scenarios. For the Stage environment evaluation, a corner turning scenario (Env2) was implemented. The environmental variation rates of DRL-NSUO and IPAPRec (a distance-based DRL algorithm) were evaluated during corner navigation. The comparative results under controlled conditions (identical target point, starting point, speed, and angular velocity) are presented in Fig. \ref{comparison chart}. During navigation tasks that involve perspective blind spots, particularly in cornering scenarios, DRL-NSUO demonstrates superior performance by maintaining minimal environmental variation rates compared to distance-based methods. This characteristic enables a more robust perception zone during turns and obstacle avoidance, thus mitigating the risks associated with sudden intrusions from dynamic obstacles or pedestrians. Performance evaluation used a simulated black rectangular-shaped mobile robot (SRob1) in the Env2 scenario. SRob1 and the test robot (SRob0) were configured to navigate along different trajectories that converge at a corner. SRob1 moves at a constant speed of 1.7m/s, simulating a cleaning robot performing tasks in the real world, and is guaranteed to encounter SRob0 at the corner. The experimental results are presented in Fig. \ref{chutian2} and Fig. \ref{chutian1}. The experimental setup compares two navigation methods: DRL-NSUO (on the left) and IPAPRec (on the right). DRL-NSUO demonstrates improved safety performance in challenging scenarios, such as closed corners, which are characterized by elevated environmental change rates, perceptual blind spots, and compromised safety conditions.

\begin{table*}[tb]
\centering
\caption{\label{tab:simulation results}Comparison of Different Methods using BARN Dataset under Two Speed Settings.}
\renewcommand\arraystretch{1.6} 
\setlength{\tabcolsep}{3mm} 

\begin{tabular}{l|cccc|cccc|c|c}
\hline
\hline
Method       & \multicolumn{4}{c|}{Speed: 0.5 m/s} & \multicolumn{4}{c|}{Speed: 1.0 m/s} & DRL-based & Use Global Path \\
             & Metric($\uparrow$) & SR ($\uparrow$) & CR ($\downarrow$) & TO ($\downarrow$) & Metric($\uparrow$) & SR ($\uparrow$) & CR ($\downarrow$) & TO ($\downarrow$) & & \\
\hline
DWA          & 0.2354              & 55.0\%          & 9.0\%            & 36.0\%              & 0.207              & 43.0\%          & 26.0\%            & 31.0\%              & No        & Yes \\
Fast-DWA     & 0.2471              & 55.6\%          & 11.1\%            & 33.3\%              & 0.2095              & 44.0\%          & 23.0\%            & 33.0\%              & No        & Yes \\
E-Band       & 0.3244              & 65.0\%          & 8.0\%             & 27.0\%              & 0.225              & 45.0\%          & 7.0\%            & 48.0\%              & No        & Yes \\
E2E          & 0.2520              & 53.0\%          & 42.0\%            & 5.0\%             & 0.3439              & 69.0\%          & 27.0\%            & 4.0\%              & Yes       & No  \\
DRL-VO       & 0.4444              & 88.9\%          & 8.1\%            & 3.0\%              & 0.4301              & 86.1\%          & 10.9\%            & \textbf{3.0\%}              & Yes       & Yes \\
\hline
\rowcolor{lightgray} \textbf{DRL-NSUO}     & \textbf{0.4685}              & \textbf{94.0\%}          & \textbf{3.0\% }            & \textbf{3.0\% }             & \textbf{0.4474}              & \textbf{91.0\%}          & \textbf{4.0\%}            & 5.0\%              & \textbf{Yes}       & \textbf{No}  \\
\hline 
\hline 
\end{tabular}

\begin{minipage}{\linewidth}
    \vspace{3pt}
    \footnotesize
    \textbf{E2E is trained using the test environments.} 
\end{minipage}
\end{table*}

\subsection{Test in Gazebo}
\begin{figure}
    \centering
    \includegraphics[width=0.8\linewidth]{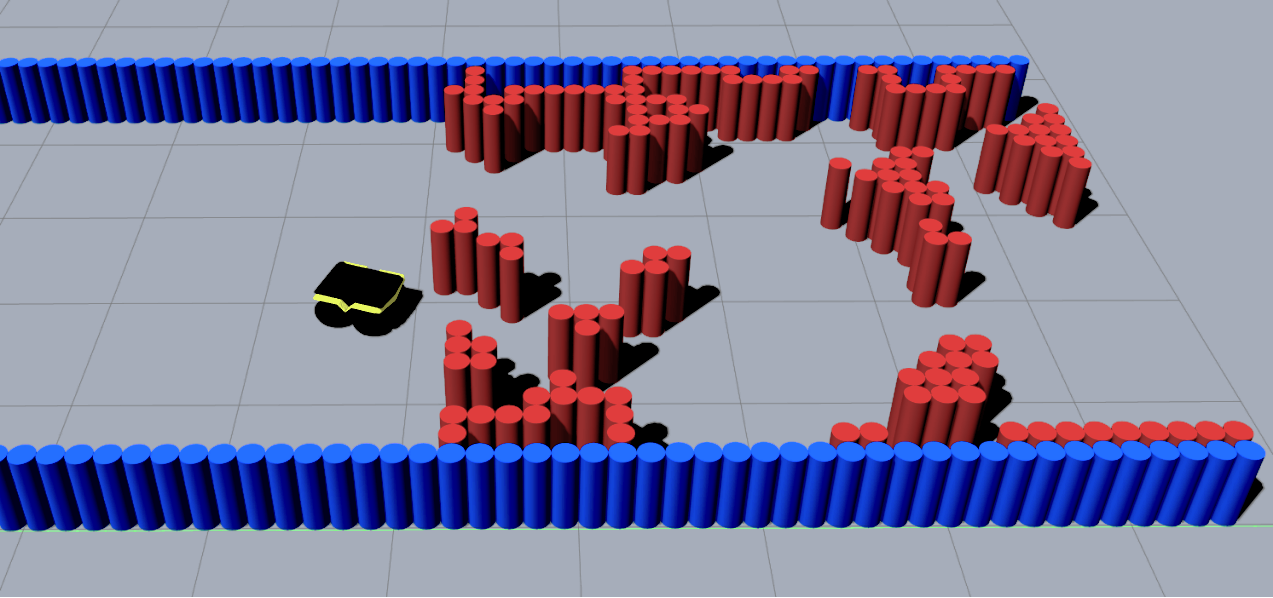}
    \caption{BARN Dateset Validation Scenario Example.}
    \label{BARN}
\end{figure}

Beyond ensuring safety, to validate the navigation of our algorithm in crowded environments, we used the Benchmark for Autonomous Robot Navigation (BARN) dataset\cite{9292572}.The BARN dataset, built on the Gazebo simulator, features complex environments like narrow corridors, densely obstructed areas, large open spaces, and dynamic obstacles. The validation process consists of 100 randomly selected test environments, each with a different obstacle configuration that represents a real world scenario. As illustrated in Fig. \ref{BARN}, the experimental platform utilized the Jackal robot, a four-wheel drive mobile platform with dimensions of 0.42 meters in length and 0.21 meters in width. The navigation model was subsequently retrained in the Ros Stage based on the Jackal robot's specifications. Performance evaluation in the BARN test environment was conducted at two distinct velocities (1 m/s and 0.5 m/s) to assess DRL-NSUO's safety and obstacle avoidance capabilities. And we specifically tested the scores of other traditional and state-of-the-art planners based on this BARN challenge dataset, including: 
\begin{itemize}
    \item Dynamic Window Approach (DWA) \cite{Fox1997TheDW}
    \item Fast Dynamic Window Approach (Fast-DWA) \cite{xiao2022autonomous}
    \item Elastic Bands (E-band) \cite{291936}
    \item End-to-End (E2E) \cite{xiao2022autonomous}
    \item DRL-VO \cite{xie2023drl}
\end{itemize}

It should be noted that the E2E algorithm was trained based on test scenarios (BARN dataset), which provideed certain advantages in testing the generalization of the algorithm. DRL-VO, as the top performing algorithm in simulation during the 2022 BARN competition, represents the state-of-the-art in mobile robot navigation algorithms. The successful navigation capabilities of DWA, Fast-DWA, E-band, and DRL-VO are fundamentally dependent on global planners for path planning support. DRL-NSUO, operating exclusively as a local planning algorithm, experiences timeout challenges that highlight the BARN challenge's substantial reliance on global planning functionalities. 

The algorithm evaluation includes several performance metrics: average success rate (SR), average collision rate (CR), and average score $s$ on 100 tests. The performance score $s_i$ for the $i$-th test iteration is quantified according to the BARN challenge criteria as follows\cite{unknown}, 
\begin{equation}
\begin{aligned}
s_i = 1^{\text{success}} \times \frac{\text{OT}_i}{\text{clip}(\text{AT}_i, 2\text{OT}_i, 8\text{OT}_i)}
\label{eq10}
\end{aligned}
\end{equation}
where the indicator  $1^{\text{success}}$  is set to  $1$ if the robot reaches the navigation goal without collisions, and 0 otherwise. $AT$ represents the actual traversal time, while $OT$ denotes the optimal traversal time as follows,
\begin{equation}
\begin{aligned}
\text{OT}_i = \frac{\text{Path Length}_i}{\text{Maximal Speed}}
\label{eq10}
\end{aligned}
\end{equation}

\begin{figure}[t]
	\centering\subfloat[DWA]{
	\centering\includegraphics[width=0.282\linewidth]{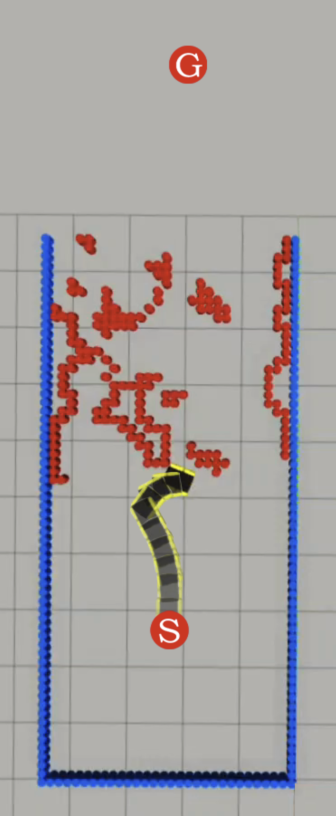}\label{t1}}
    \hspace{0.02\linewidth}
	\subfloat[E2E]{
    \centering\includegraphics[width=0.277\linewidth]{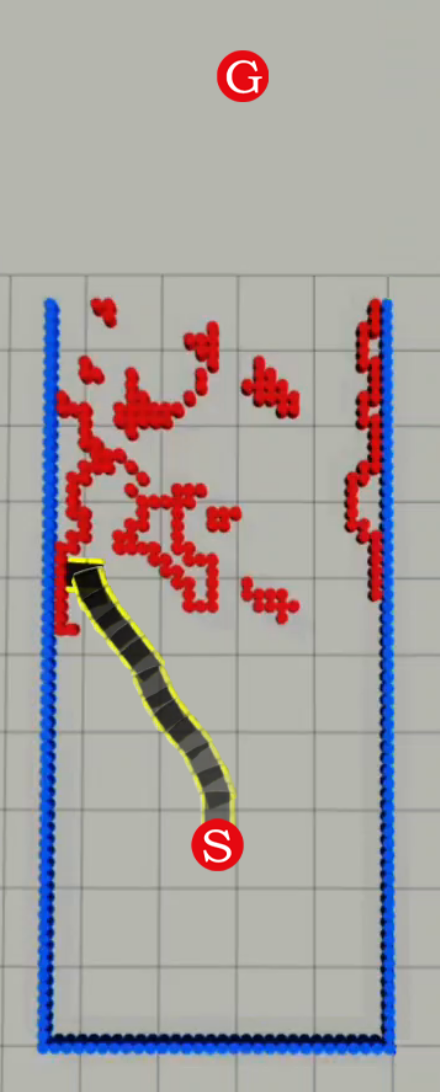}\label{t2}}
    \hspace{0.02\linewidth}
    \subfloat[DRL-NSUO]{	\centering\includegraphics[width=0.284\linewidth]{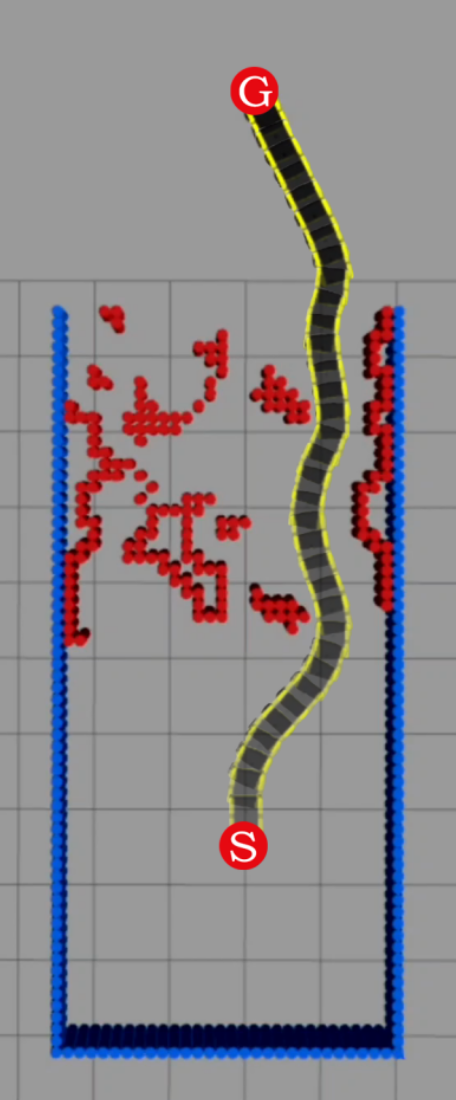}\label{t3}}
    
	\caption{Trajectories of the robot controlled by (a) DWA, (b) E2E, and (c) DRL-NSUO in No.207 BARN testing map.}
	\label{sim_compare}
\end{figure}

The test results are summarized in Table 1, DWA and DRL-VO represent traditional methods and the current state-of-the-art deep reinforcement learning navigation methods, respectively. Compared to DWA, DRL-NSUO significantly outperformed DWA with a Metric of 0.4685, success rate of 94\%, and collision rate of 3\% at 0.5 m/s, while DWA's values were 0.2354, 55\%, and 9\%, respectively. At 1.0 m/s, DRL-NSUO continueed to outperform with a Metric of 0.4474, success rate of 91\%, and collision rate of 4\%, while DWA's performance significantly declined. Compared to the DRL-VO and E2E methods, DRL-NSUO showed a more stable performance at both speeds, particularly in terms of collision rate and time-out rate, where the advantages of DRL-NSUO were even more pronounced. 
As shown in Fig. \ref{sim_compare}, DRL-NSUO showed excellent obstacle-avoidance navigation performance on complex maps. 
More importantly, DRL-NSUO operates without reliance on global path planning methods, demonstrating superior adaptability and reliability in local decision-making while maintaining excellent performance metrics. Consequently, this method achieved the highest ranking among the evaluated approaches, offering enhanced flexibility and reduced operational risk compared to alternative navigation strategies.

\begin{figure*}[t]
	\centering\subfloat[REnv1]{
	\centering\includegraphics[width=0.17\linewidth]{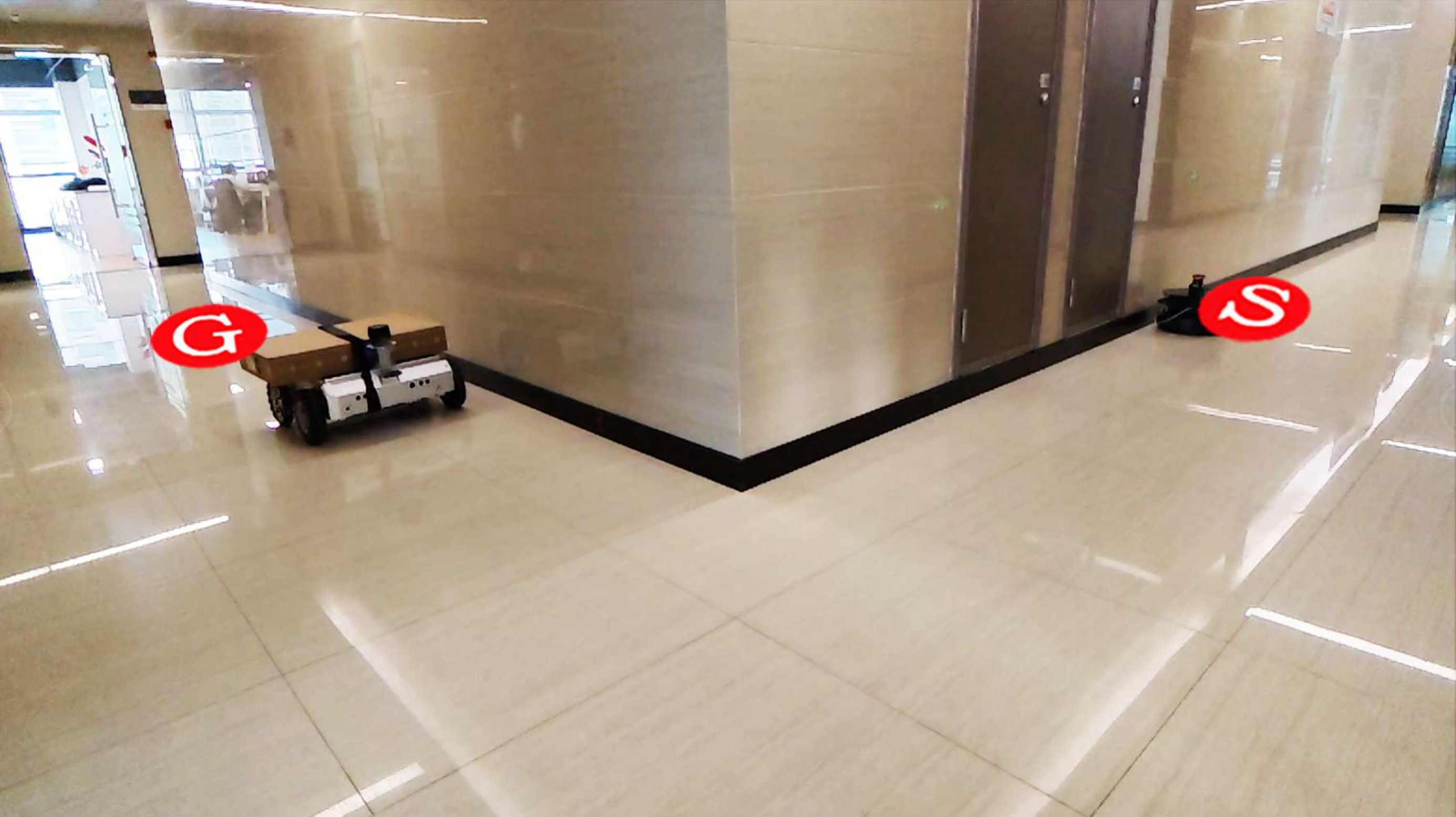}\label{t1}}
	\hfill\subfloat[REnv2]{
	\centering\includegraphics[width=0.17\linewidth]{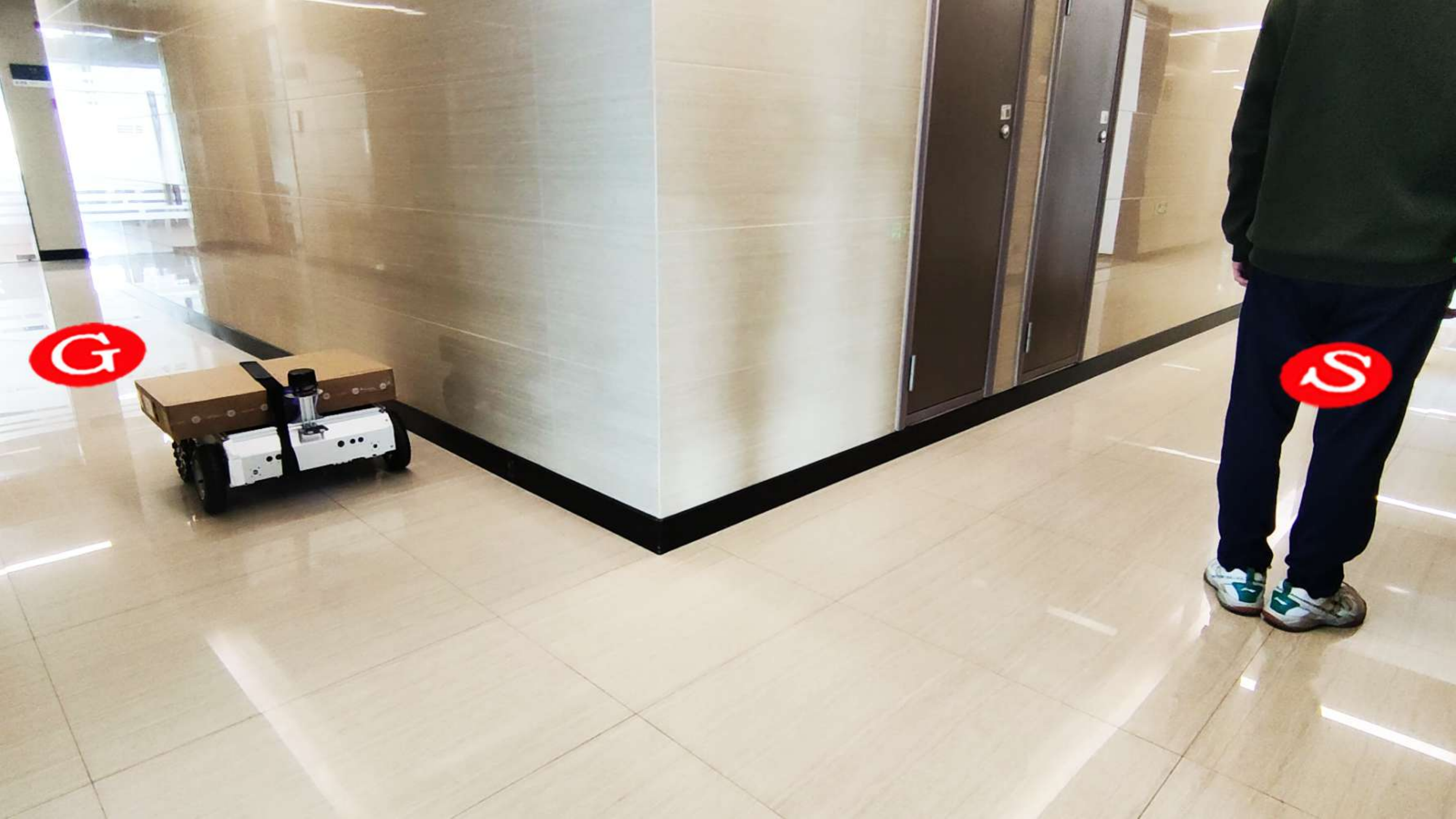}\label{t2}}
	\subfloat[REnv3]{
	\centering\includegraphics[width=0.169\linewidth]{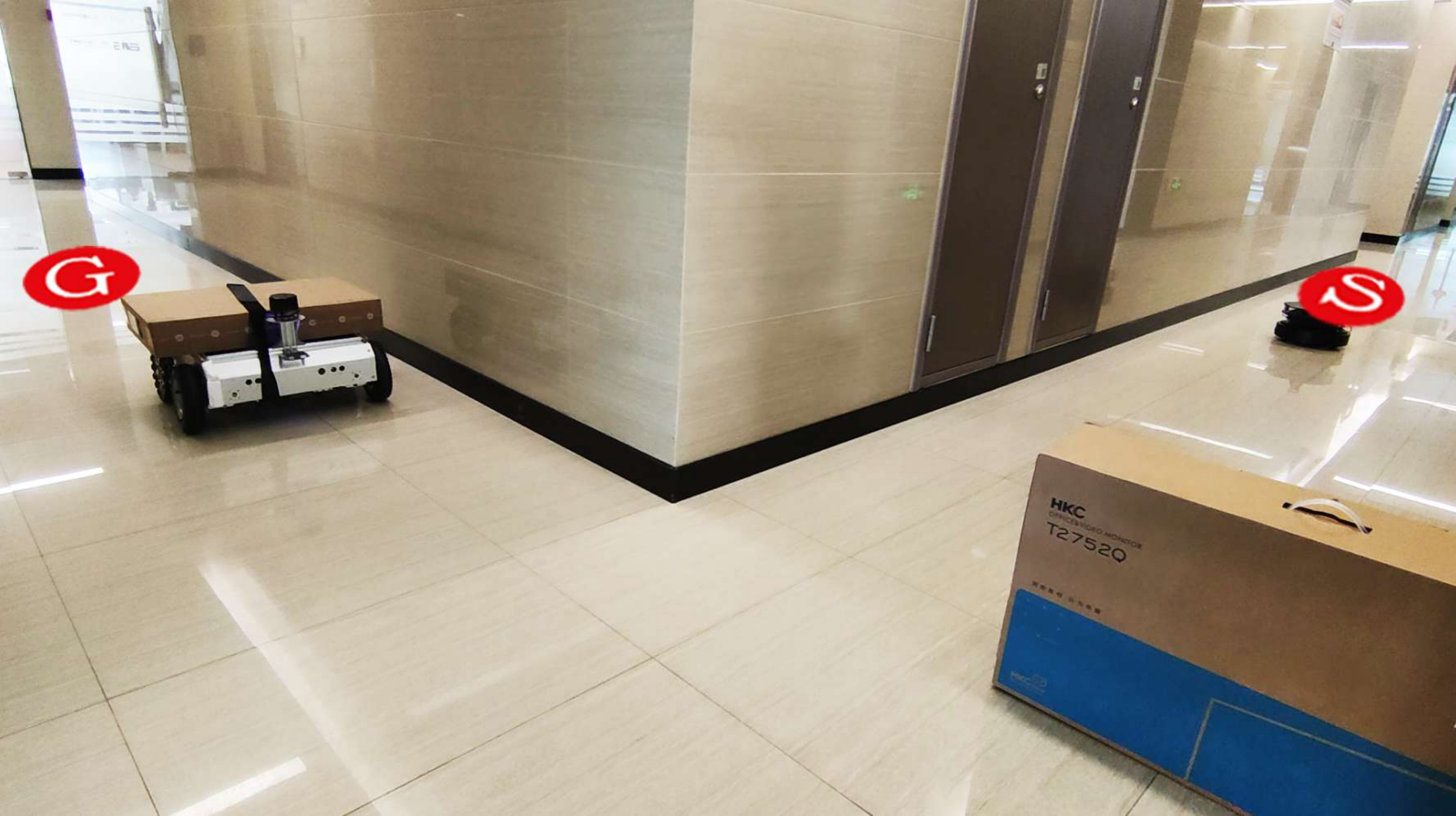}\label{t3}}
	\hfill\subfloat[REnv4]{
	\centering\includegraphics[width=0.167\linewidth]{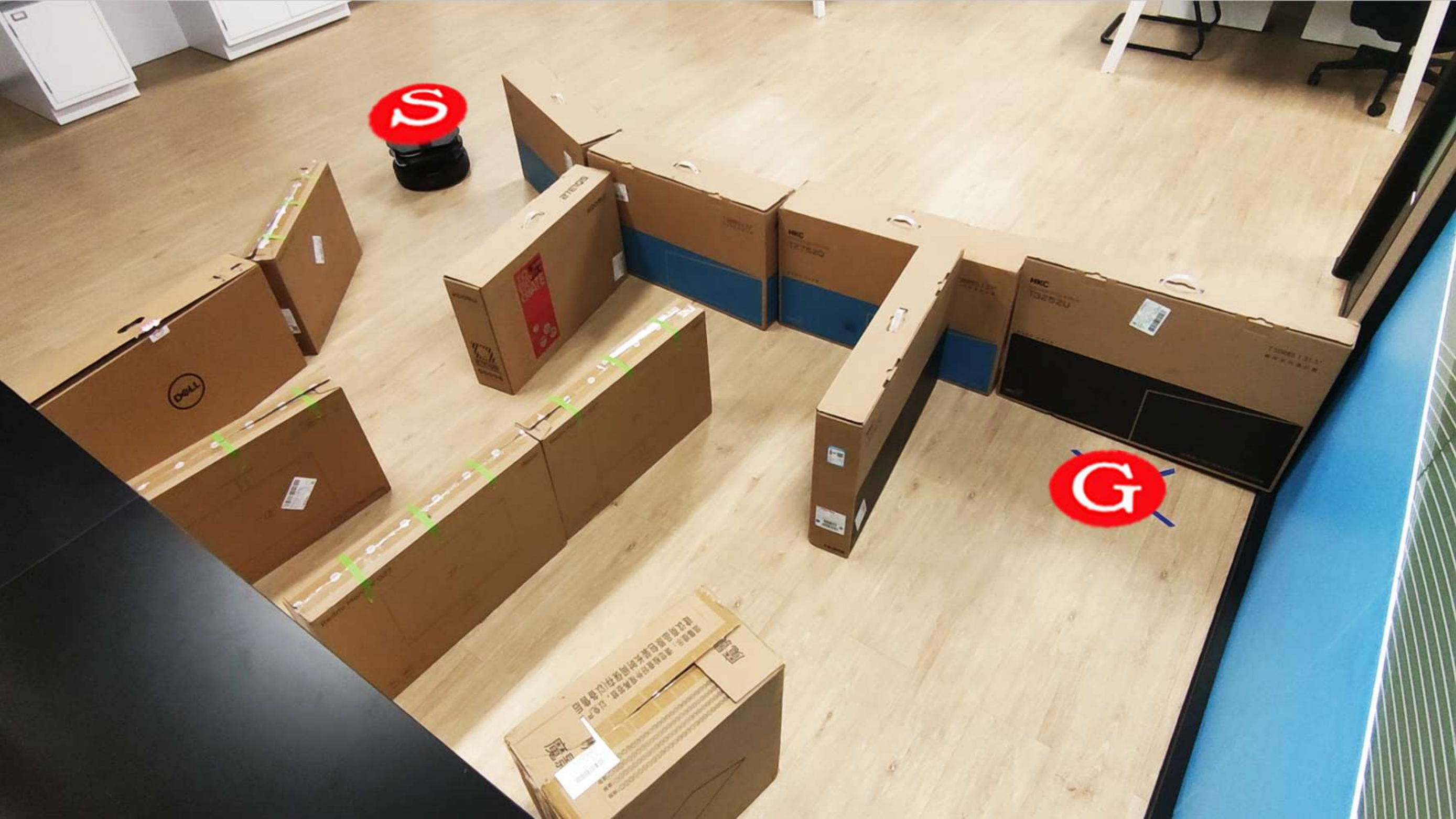}\label{t4}}
    \hfill\subfloat[REnv5]{
	\centering\includegraphics[width=0.165\linewidth]{che.pdf}\label{The robot}}	
    \hfill\subfloat[Test Robot]{
	\centering\includegraphics[width=0.1\linewidth]{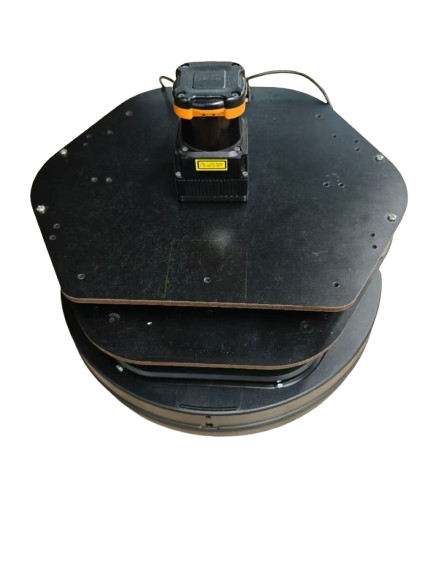}\label{t1}}

	\caption{Real-world experimental setup showing test environments (a-e) and the mobile robot platform (f) used for validation. Start points are marked as `S' and target points as `G'.}

	\label{real_test_env}
\end{figure*}

\begin{figure}[t]
	\centering\subfloat[DRL-NSUO in REnv1]{
	\centering\includegraphics[width=0.48\linewidth]{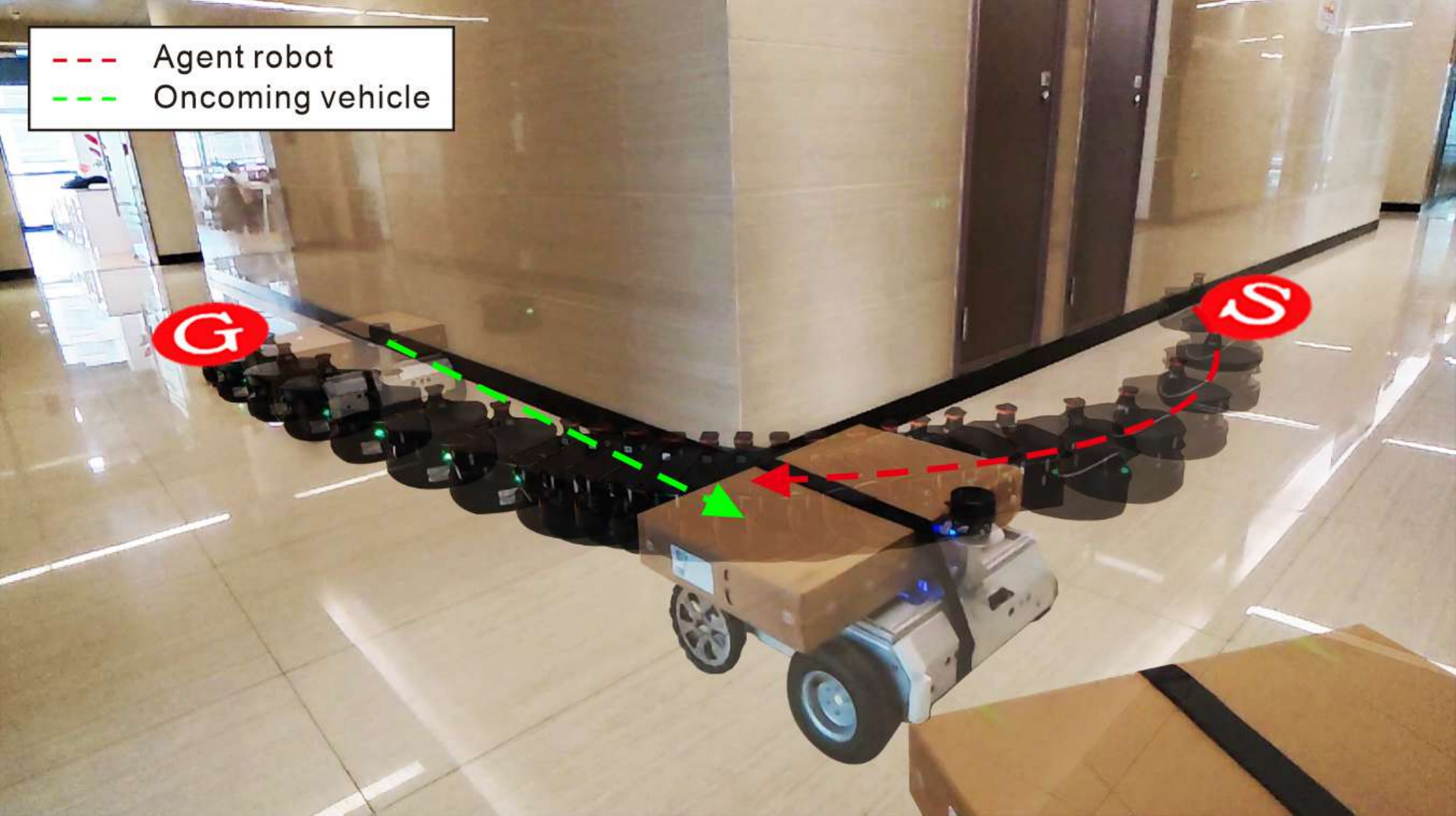}\label{t11}}
	\hfill\subfloat[IPAPRec in REnv1]{
	\centering\includegraphics[width=0.48\linewidth]{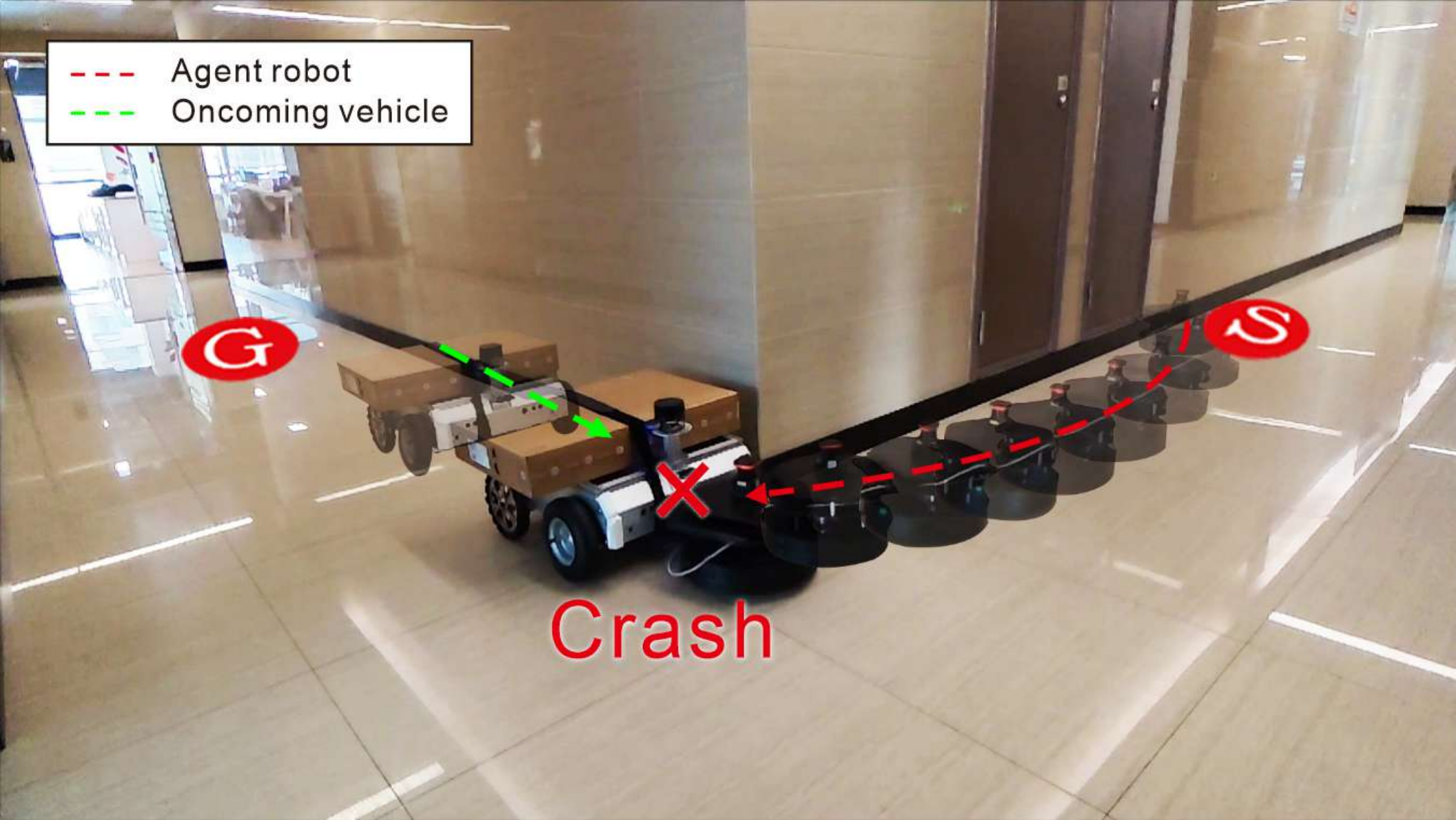}\label{t21}}
	\\[-0ex]
	\subfloat[DRL-NSUO in REnv3]{
	\centering\includegraphics[width=0.48\linewidth]{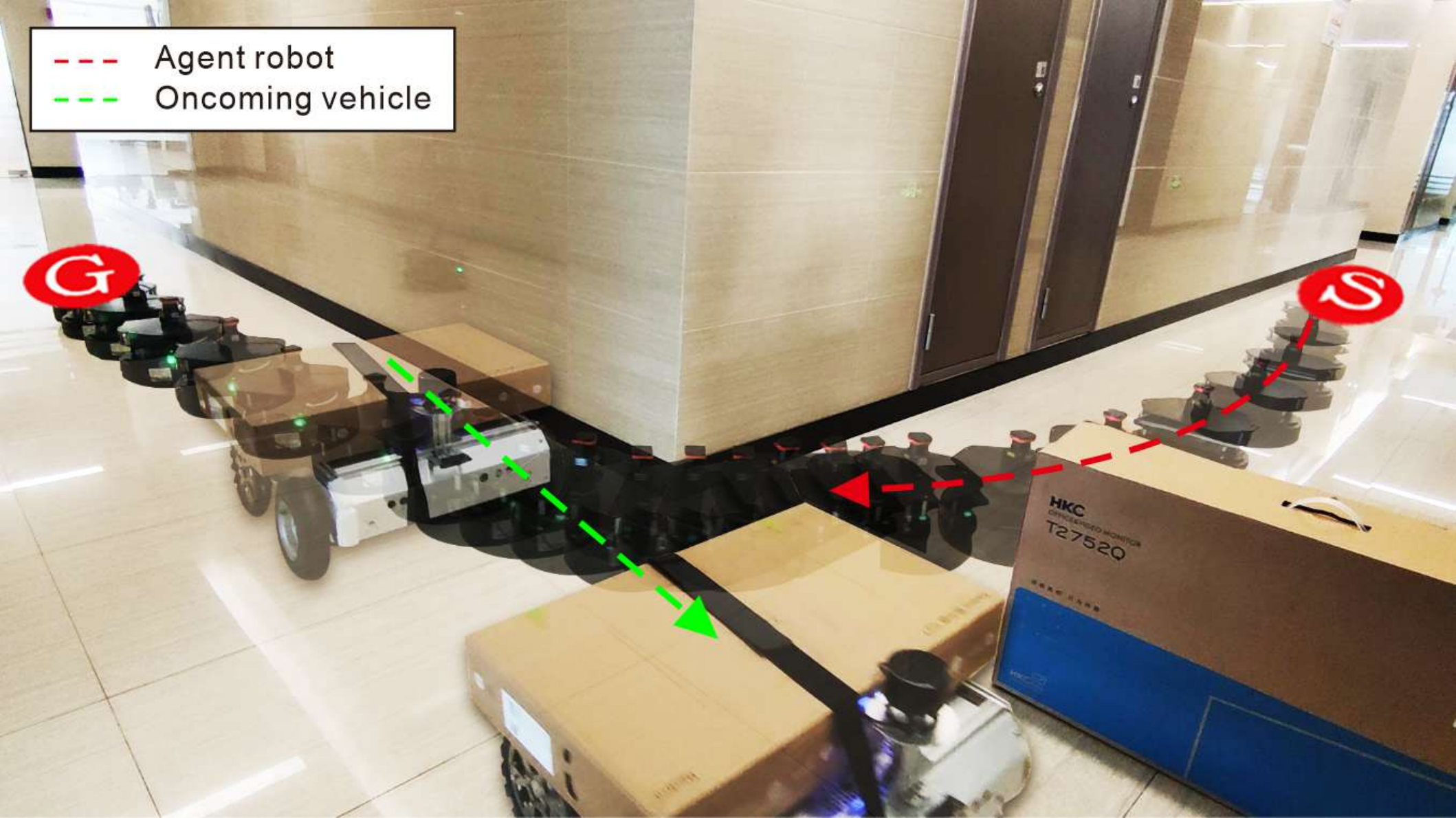}\label{t31}}
	\hfill\subfloat[IPAPRec in REnv3]{
	\centering\includegraphics[width=0.48\linewidth]{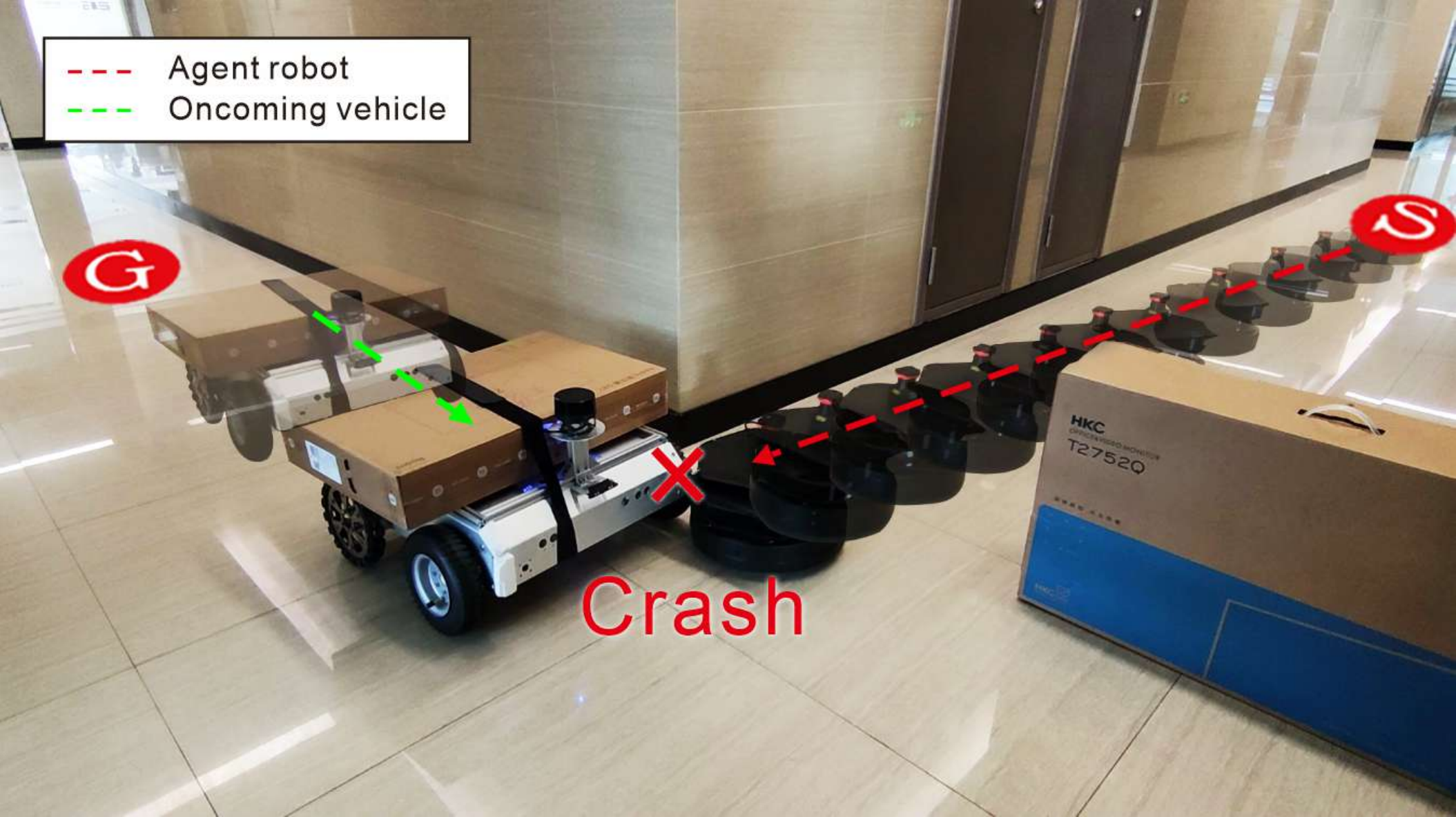}\label{t41}}
	\caption{Trajectories of robot trained with DRL-NSUO and trained with IPAPRec for comparison when tested in REnv1 and REnv3. The experimental videos can be found in the supplementary file.}
	\label{real_test}
\end{figure}

\section{Performance Validation in REAL-WORLD}
Real-world experiments validate the algorithm's outstanding safety, obstacle avoidance, and navigation performance in crowded environments.

\subsection{Hardware setup}
As illustrated in Fig. \ref{The robot}, the experimental platform consists of a Turtlebot2 robot system integrated with a Hokuyo UTM-30LX Lidar sensor. The LiDAR system features a 270-degree field of view (FOV), with specifications including a 30-meter maximum measurement range and 0.25-degree angular resolution. Given DRL-NSUO's implementation as a local planner, the system incorporates \textit{ROS Gmapping} \cite{Gmapping} for map construction and \textit{ROS AMCL} \cite{AMCL} for precise robot localization. The local planning computations were executed on an onboard laptop equipped with an i7-7600U CPU. The generated environmental map serves exclusively for robot and target point localization purposes, independent of the motion planning process.


\subsection{Test Scenarios and Task Description}
The robot was tested in five real-world scenarios, as shown in Fig. \ref{real_test_env}. These test scenarios included three corridor environments and two crowded environments. In each environment, the robot started from an initial position marked `S' and was tasked with navigating to a destination point labeled `G'. REnv1 is a corner scenario where the experimental robot needs to avoid a moving square robot obstacle going straight. REnv2 extends the REnv1 configuration by incorporating an additional dynamic element: a pedestrian that traverses in the opposite direction relative to the experimental platform. This scenario necessitates simultaneous avoidance of multiple dynamic obstacles: the pedestrian and the square robot. REnv3 augments the REnv1 scenario with a static obstacle (cardboard box) positioned at the corner. This setup requires the platform to navigate a narrow passage between a moving obstacle and a static obstacle.  REnv4 and REnv5 simulate crowded spaces, with REnv5 involving a multiobjective task of sequential target acquisition.

\begin{figure}[t]
	\centering\subfloat[DRL-NSUO in REnv2]{
	\centering\includegraphics[width=0.48\linewidth]{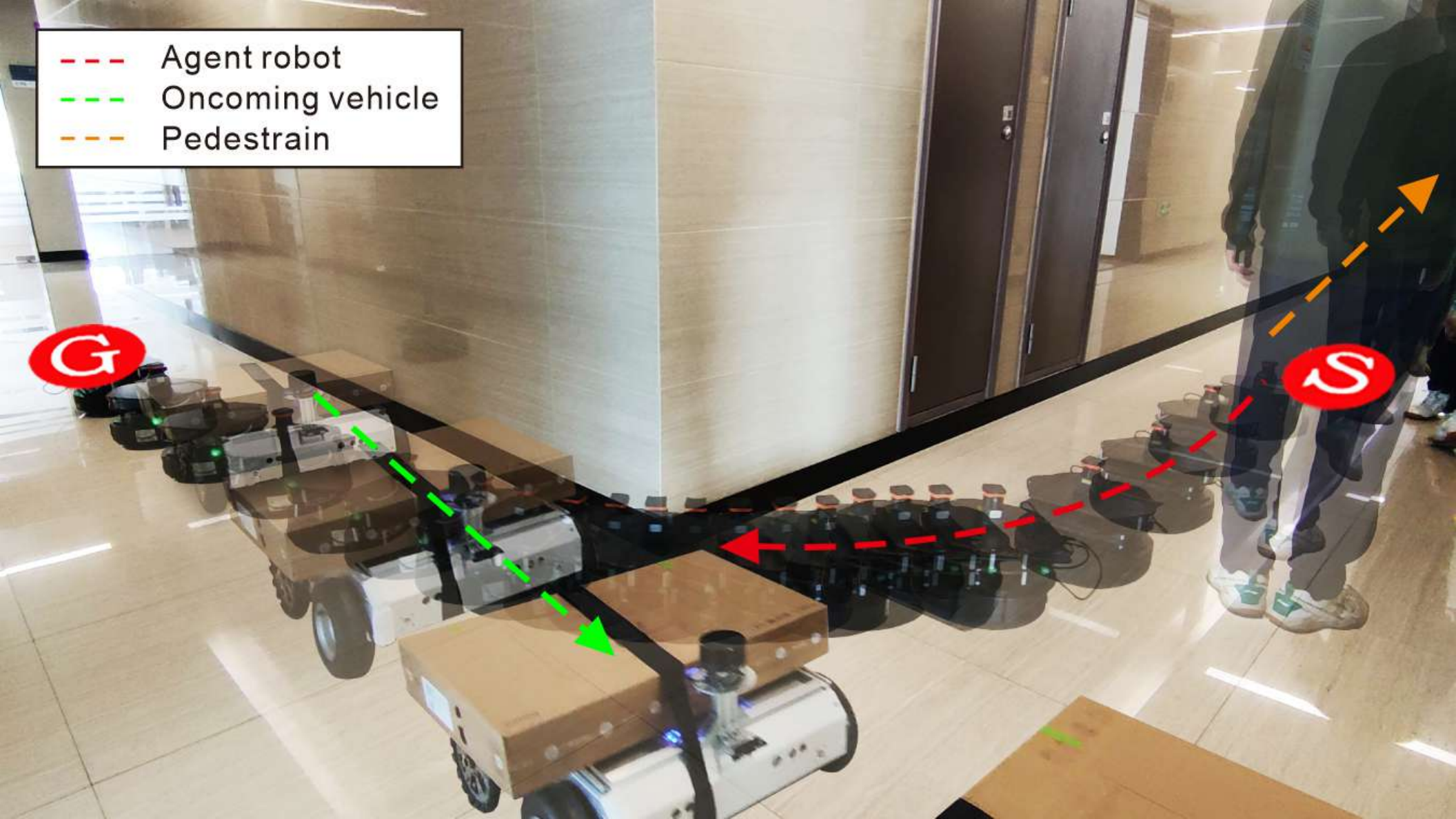}\label{ningbo1}}
	\hfill\subfloat[DRL-NSUO in REnv4]{
	\centering\includegraphics[width=0.48\linewidth]{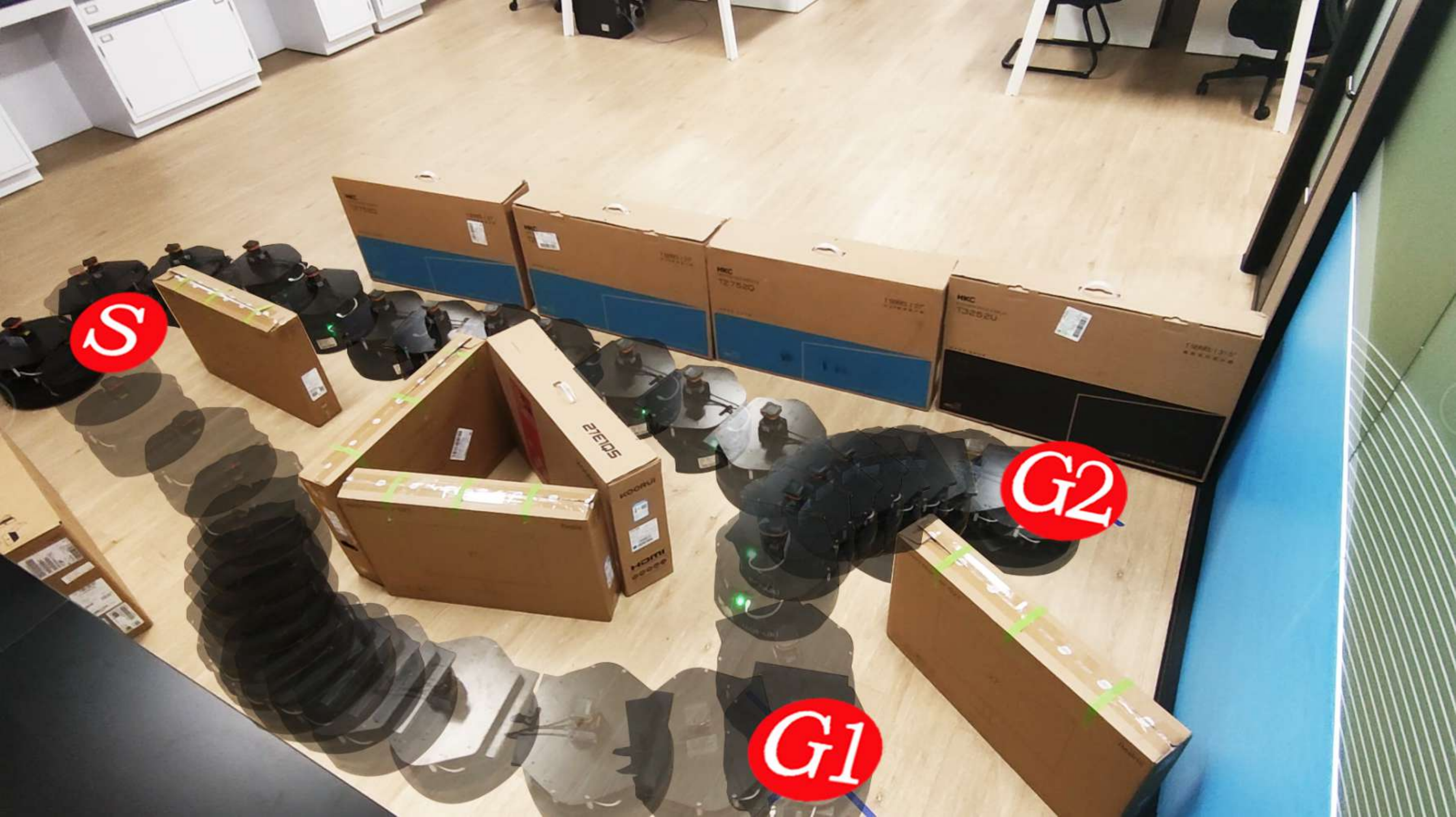}\label{ningbo2}}
	\\[-0ex]
	\subfloat[REnv4 with Pedestrian Block]{
	\centering\includegraphics[width=0.48\linewidth]{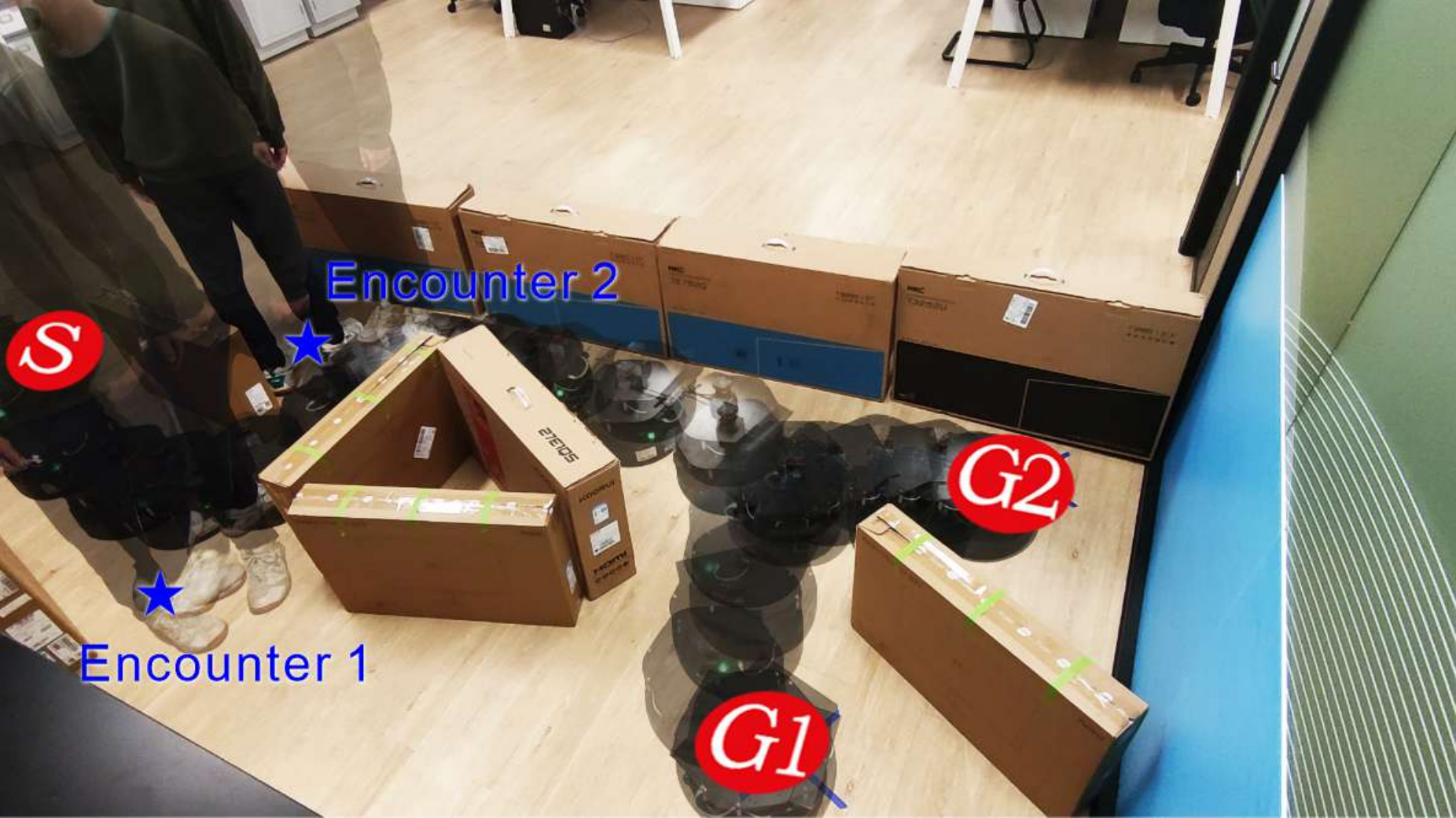}\label{ningbo3}}
	\hfill\subfloat[DRL-NSUO in REnv5]{
	\centering\includegraphics[width=0.48\linewidth]{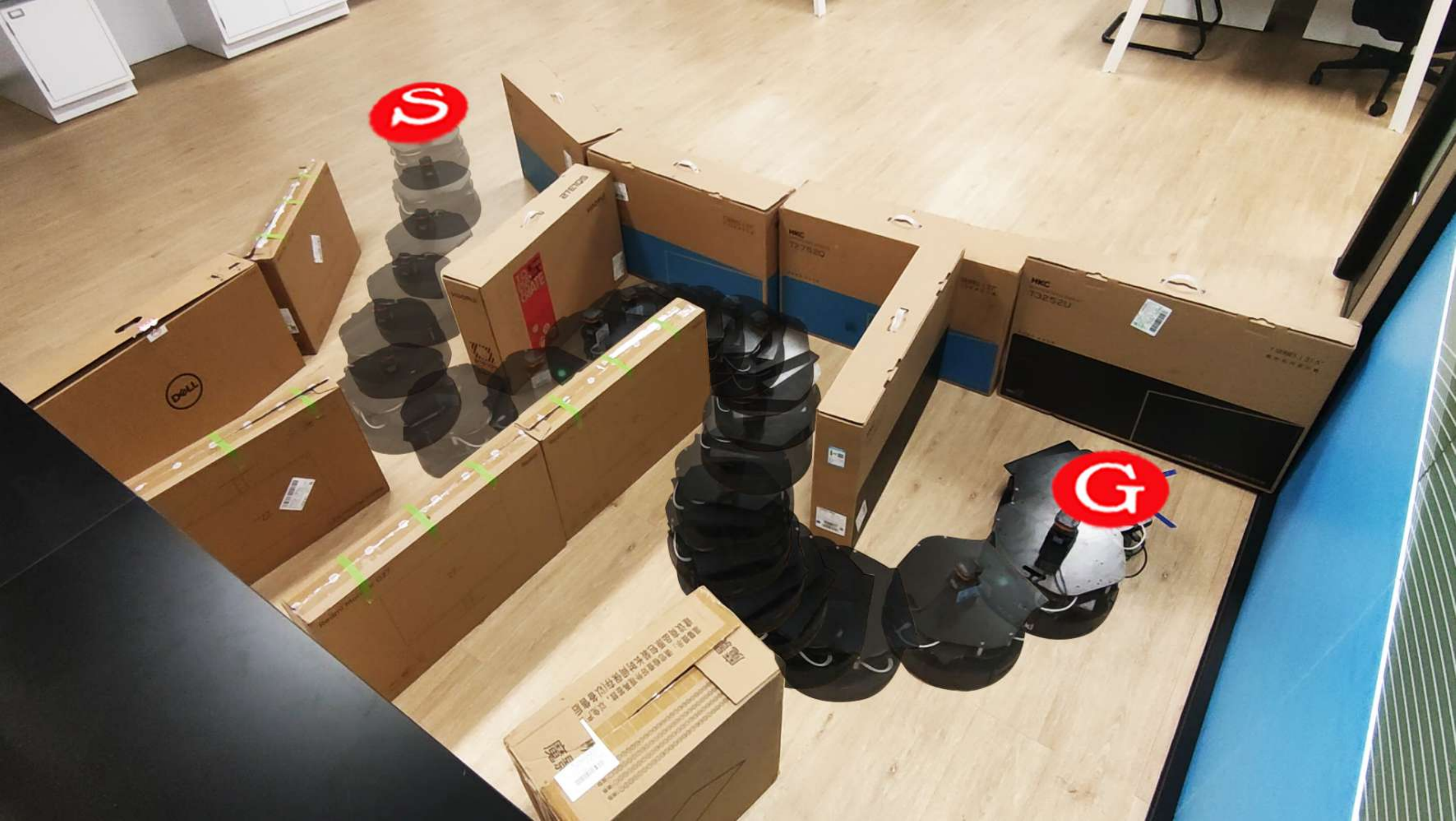}\label{ningbo4}}
	\caption{Trajectories of robot trained with DRL-NSUO when tested in the REnv2, REnv4 and REnv5. The experimental videos can be found in the supplementary file.}
	\label{real_test_ningbo}
\end{figure}

\subsection{Experiment Result}
The DRL-NSUO algorithm demonstrated superior performance compared to conventional distance-based DRL navigation algorithms, exhibiting enhanced safety characteristics while maintaining robust obstacle avoidance and navigation capabilities in crowded environments. The comparative analysis of DRL-NSUO and IPAPRec methodologies in environments REnv1 and REnv3 is presented in Fig. \ref{real_test}. Each subplot illustrates the robot's navigational trajectory throughout the experimental trial. In the REnv1 environment (Fig. \ref{t11} and Fig. \ref{t21}), the DRL-NSUO methodology demonstrated adaptive velocity control at corners while maintaining minimal environmental change rates at intersections, facilitating optimal trajectory planning toward road centers and successful dynamic obstacle avoidance. Conversely, the IPAPRec method exhibited significant limitations in dynamic obstacle handling, resulting in a collision with the mobile robot during high-speed corner navigation. In the REnv3 scenario (Fig. \ref{t31} and Fig. \ref{t41}), characterized by corner navigation constrained due to a large static obstacle, DRL-NSUO demonstrated effective velocity modulation, allowing successful obstacle avoidance. The empirical results demonstrated DRL-NSUO's superior robustness and enhanced responsiveness to complex, dynamic environmental variations. Conversely, IPAPRec exhibited reduced environmental adaptability, resulting in increased vulnerability to environmental variations and subsequent task failures.Fig. \ref{ningbo1} demonstrated successful navigation in REnv2, featuring simultaneous avoidance of pedestrians and dynamic obstacles while maintaining goal-directed behavior. The navigation performance in REnv4 (Fig. \ref{ningbo2}) demonstrated the successful arrival at sequential goals (G1 and G2), even with the increased environmental complexity. Fig. \ref{ningbo3} showed the effectiveness of multi-pedestrian avoidance in REnv4, where the robot successfully avoided pedestrians, autonomously adjusted the return trajectory, and smoothly reached all the target points. Fig. \ref{ningbo4} presented the navigation trajectory in REnv5, characterized by elevated environmental complexity and dense obstacle distribution. The system demonstrated robust performance in this challenging scenario, exhibiting effective collision avoidance, appropriate velocity modulation at corners, and smooth access to the target point.

\section{Conclusion}
In this work, we present DRL-NSUO, a novel DRL-based navigation strategy specifically designed to handle unexpected obstacles. The implementation of a reward mechanism based on environmental change rate significantly improves robotic safety performance in dynamic and unstructured environments. The developed reward mechanism demonstrates dual functionality: optimizing distance-based path planning while simultaneously mitigating the effects of dynamic environmental perturbations through continuous minimization of environmental change rates. The proposed methodology maintains robust obstacle avoidance capabilities while substantially enhancing navigation safety in complex, crowded, and dynamic environments. Quantitative evaluation using the BARN navigation dataset demonstrated DRL-NSUO's superior performance, achieving optimal success rates and metric scores. Experimental validation confirmed the enhanced efficacy of DRL-NSUO in complex environment navigation compared to conventional deep-reinforcement learning and path efficiency methodologies. The developed DRL-NSUO methodology presented a robust and practical solution for autonomous robot navigation in challenging real-world environments, offering significant advantages for practical applications.

\bibliographystyle{IEEEtran} 
\bibliography{mybib} 

\end{document}